\documentclass[10pt,twocolumn,a4]{article}

\usepackage{cvpr}
\usepackage{times}
\usepackage{epsfig}
\usepackage{graphicx}
\usepackage{amsmath}
\usepackage{amssymb}

\usepackage{booktabs}
\usepackage{multirow}
\usepackage{multicol}
\usepackage{array}
\newcolumntype{x}[1]{>{\centering\arraybackslash\hspace{0pt}}p{#1}}
\usepackage{comment}
\usepackage{gensymb}
\usepackage{float}
\usepackage{stfloats}
\usepackage{caption}
\usepackage{enumitem}
\usepackage{color}
\usepackage{subcaption}
\usepackage{placeins}
\usepackage[accsupp]{axessibility}
\usepackage[pagebackref=true,breaklinks=true,letterpaper=true,colorlinks,bookmarks=false]{hyperref}

\cvprfinalcopy % *** Uncomment this line for the final submission

\def\cvprPaperID{5459} % *** Enter the CVPR Paper ID here

\begin{document}

%%%%%%%%% TITLE
\title{CrossLoc: Scalable Aerial Localization Assisted by Multimodal Synthetic Data}

% \and

\author{
Qi Yan, Jianhao Zheng, Simon Reding, Shanci Li, Iordan Doytchinov\\
École Polytechnique Fédérale de Lausanne (EPFL) - TOPO laboratory\\
{\tt\small\{firstname.lastname\}@epfl.ch}
}

% \maketitle

%\thispagestyle{empty}

%%%%%%%%% Pull figure
\twocolumn[{%
\renewcommand\twocolumn[1][]{#1}%
\maketitle
\vspace{-2.5em}
\centering
\href{https://crossloc.github.io}{\url{crossloc.github.io}}
\begin{center}
    \centering
    \captionsetup{type=figure}
    \includegraphics[width=.99\textwidth]{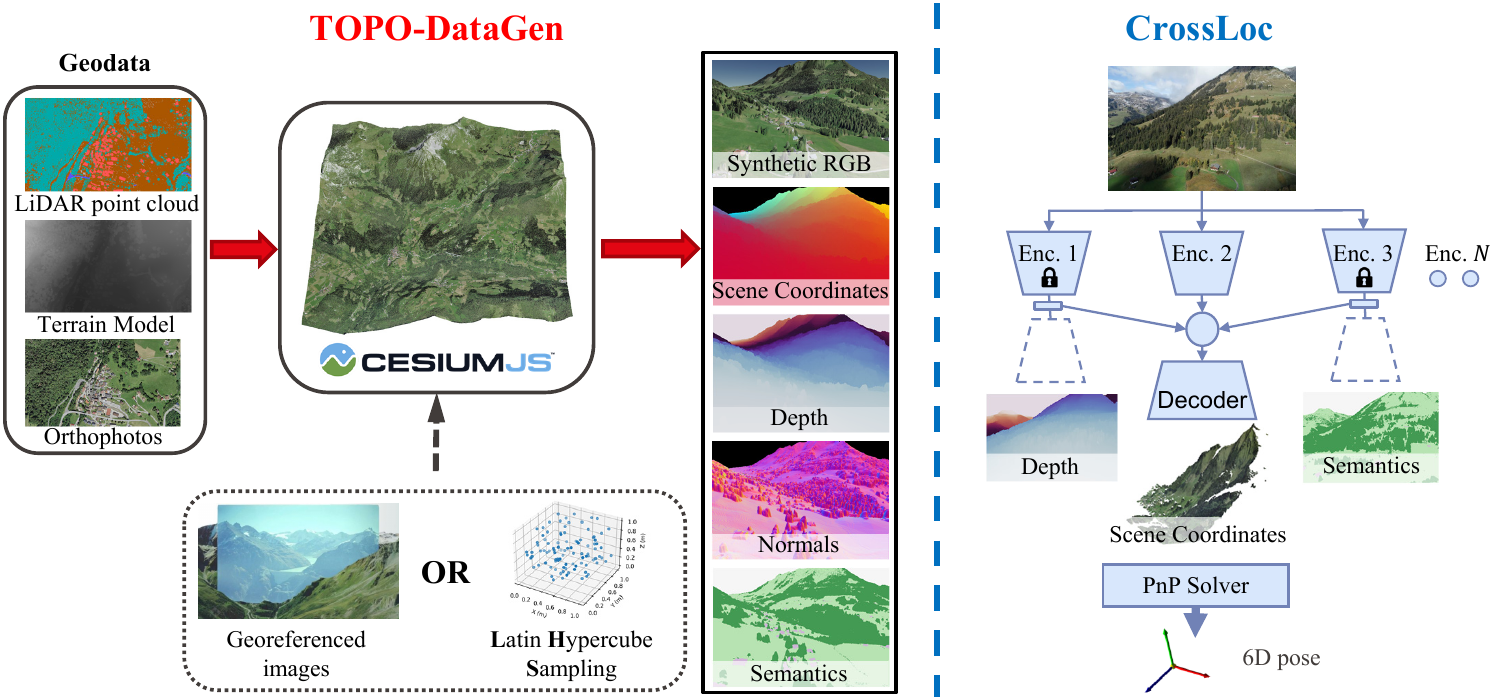}
    \captionof{figure}{
    % \textbf{System overview.} 
    \textbf{Left.}
    The proposed \textit{TOPO-DataGen} workflow generates multimodal synthetic datasets over large scales using off-the-shelf geodata.
    \textbf{Right.}
    We present \textit{CrossLoc}, a cross-modal visual representation learning method for absolute localization.
    It learns to predict scene coordinates via self-supervised geometric tasks and external labels such as semantics if available.
    % It exemplifies the utility of \textit{TOPO-DataGen} .
    }
	\label{Figures:main}
\end{center}%
}]

%%%%%%%%% ABSTRACT
\begin{abstract}
We present a visual localization system that learns to estimate camera poses in the real world with the help of synthetic data.
Despite significant progress in recent years, most learning-based approaches to visual localization target at a single domain and require a dense database of geo-tagged images to function well.
To mitigate the data scarcity issue and improve the scalability of the neural localization models, we introduce TOPO-DataGen, a versatile synthetic data generation tool that traverses smoothly between the real and virtual world, hinged on the geographic camera viewpoint.
New large-scale sim-to-real benchmark datasets are proposed to showcase and evaluate the utility of the said synthetic data.
Our experiments reveal that synthetic data generically enhances the neural network performance on real data.
Furthermore, we introduce CrossLoc, a cross-modal visual representation learning approach to pose estimation that makes full use of the scene coordinate ground truth via self-supervision.
Without any extra data, CrossLoc significantly outperforms the state-of-the-art methods and achieves substantially higher real-data sample efficiency.
% Our code is available at \url{https://github.com/TOPO-EPFL/CrossLoc}.
Our code and datasets are all available at
\href{https://crossloc.github.io}{\url{crossloc.github.io}}
% \url{https://github.com/TOPO-EPFL/CrossLoc}
.

\end{abstract}

%%%%%%%%% BODY TEXT

\section{Introduction}
Due to vulnerabilities in the reception of satellite positioning (GNSS) signals, on which aerial systems rely for navigation and controls, alternative methods are in demand for absolute large-scale localization. 
The dead-reckoning navigation systems have improved significantly in recent years; however, residual drift is always a challenge for long-term applications~\cite{ellingson2020relative,khaghani2018assessment}. 
In contrast, the availability of small and low-cost cameras has made them popular sensors for capturing information on the surrounding landscape. 
When features of a known environment are recognized in the captured images, these can be used to determine the absolute camera poses.
Current state-of-the-art machine-learning-based methods for absolute localization~\cite{brachmann2017dsac, brachmann2019expert, brachmann2021visual, sarlin2019coarse, sarlin2021back, toft2020longterm} show promising performance but typically focus on single-domain operation and on bespoke datasets collected for indoor or outdoor city street localization~\cite{meyer2020longterm}. 
However, open-source datasets for positioning airborne platforms or workflows to generate geospatial learning data are scarce. 
This poses a severe barrier in adapting the algorithms to real-world aerial scenarios, as they typically require dense datasets consisting of accurately geo-referenced photos in the area of interest~\cite{ jafarzadeh2021crowddriven}. 
Developing inclusive datasets for real-life navigation around the globe and employing state-of-the-art visual algorithms for aerial applications can currently be considered economically and technically difficult, if not unfeasible. 
\par
In this work, we first present a synthetic data generation scheme called \textbf{TOPO-DataGen} (Figure~\ref{Figures:main}, left), that leverages the topographic information to produce geo-referenced data with rich modalities for subsequent training. 
Given the designated camera poses, this scheme renders the simulated RGB images accompanied by 2D and 3D modalities such as semantics, scene coordinates, depth, and surface normal.
In an area of interest with available geodata, one may adopt a stochastic sampling strategy such as Latin hypercube sampling (LHS)~\cite{Florian1992AnSampling,Manteufel2000EvaluatingSampling} or use real geo-tagged photos to attain the camera viewpoints and create the corresponding synthetic labels.
% via TOPO-DataGen.
The real geo-tagged data can be designer sourced, such as data acquisition by drones or from crowd sourced campaigns~\cite{mapillary, smapshot}.
% such as the Smapshot~\cite{smapshot} or mapillary~\cite{mapillary}. 
Our method hinges on the geospatial location of the camera viewpoint and traverses between reality and simulation smoothly.
% , $e.g.,$ one can intentionally create the matching 3D cues for the geo-referenced real data at hand.

To mitigate the data-scarcity issue for learning-based visual localization methods via sim-to-real transfer, we curated two large-scale \textbf{benchmark datasets} at~\cite{iordan2022crossloc} using the proposed data generation workflow on urban and natural sites.
% \footnote{https://github.com/TOPO-EPFL/TOPO-DataGen}
They are comprised of primarily synthetic data and a small fraction of accurately geo-tagged real data, with both sections containing dense 3D and semantic labels.
Unlike the existing datasets focusing on localization in a single domain~\cite{jafarzadeh2021crowddriven, kendall2015posenet, sattler2018benchmarking}, the provided benchmark datasets showcase and evaluate the use of synthetic data to assist localization in the real world using significantly less real data.

In addition, we introduce a cross-modal visual representation learning approach \textbf{CrossLoc} (Figure~\ref{Figures:main}, right) for absolute localization via scene coordinate regression.
CrossLoc exploits the rich information contained in the scene coordinates through self-supervision to achieve improved performance.
We start from the scene coordinate ground-truth to impose geometrically less complex pretext tasks such as depth estimation without any extra labels. 
The visual representations learned from the tightly-coupled tasks~\cite{qi2018geonet, zamir2020robust, zamir2018taskonomy} jointly improve the downstream coordinate regression.
We find that this approach consistently outperforms the state-of-the-art baselines in our benchmark.
\par  

Our main contributions are summarized as follows:
\begin{enumerate}[topsep=0pt,itemsep=-1ex,partopsep=1ex,parsep=1ex]
   \item TOPO-DataGen: an open-source multimodal synthetic data generation tool tailored to aerial scenes.
%   \item TOPO-DataGen: an open-source multimodal synthetic data generation tool scalable to anywhere for which topographic data is available.
   \item Large-scale benchmark datasets for sim-to-real visual localization, including synthetic and real images with 3D and semantic labels on urban and natural sites.
   \item CrossLoc: a cross-modal visual representation learning method via self-supervision for absolute localization, which outperforms the state-of-the-art baselines.
\end{enumerate}
% Our code and datasets are all available at \url{https://github.com/TOPO-EPFL/CrossLoc}.
%------------------------------------------------------------------------
\section{Related work}
\begin{figure*}
    \centering
    \includegraphics[width=.99\textwidth]{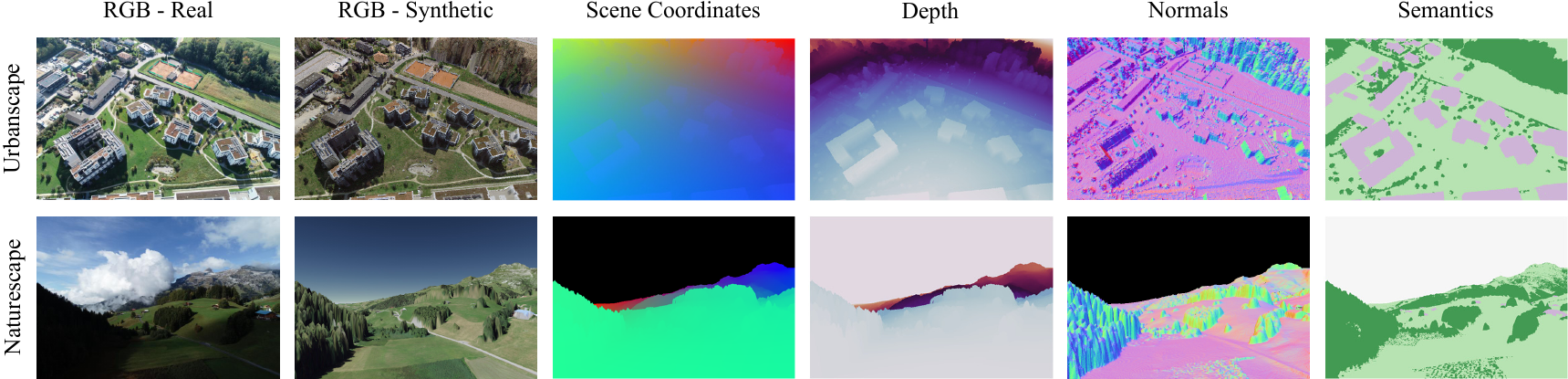}
    \caption{Examples of synthetic modalities generated with TOPO-DataGen.
    Our workflow traverses between the real and virtual world seamlessly hinged on the camera pose: from the leftmost real geo-tagged photos, we can generate a rich set of matching synthetic modalities, including synthetic RGB image, scene coordinates, depth, surface normal, and semantics.} 
	\label{Figures:Fig2}
\end{figure*}
\noindent
\textbf{Absolute visual localization} aims at estimating the camera pose of a query image within a known environment.
Many attempts have been made to achieve this using purely 2D images.
Image retrieval~\cite{arandjelovic2016netvlad, arandjelovic2014dislocation, sattler2016largescale, torii201524, torii2019are} methods rely on the image features to build an explicit localization map.
The query image is first matched against a database of reference images with known poses, and its pose w.r.t. to the retrieved image is subsequently refined via approximation or relative localization.
Absolute pose regression~\cite{brahmbhatt2018geometryaware, kendall2016modelling, kendall2015posenet, wang2020atloc} approaches adopt a neural network to learn implicit map representations and output the camera pose directly.
% Pose regression methods are proven to have similar intrinsic mechanism to image retrieval~\cite{sattler2019understanding}.
% While both are generally efficient, they fail to deliver competitive performance compared with structure-based algorithms~\cite{jafarzadeh2021crowddriven, sarlin2021back}.

Structure-based methods aim at identifying the 2D-3D matches between the query image pixels and the 3D coordinates of the scene model~\cite{brachmann2017dsac, brachmann2018learning, brachmann2019expert, brachmann2019neuralguided, brachmann2021visual, sarlin2019coarse, sarlin2021back, sattler2016efficient, toft2020longterm, yang2019sanet, zhang2021reference}.
The camera pose can then be computed using a PnP solver with RANSAC optimization~\cite{brachmann2021visual, gao2003complete,lepetit2009epnp}.
Compared with the 2D image-based approaches, the structure-based methods deliver more competitive performance~\cite{jafarzadeh2021crowddriven, sarlin2021back}, but at the cost of attaining accurate 3D models for the underlying scenes, which is non-trivial and may require specialized engineering efforts~\cite{sarlin2021back}.
The conventional descriptor matching methods, such as Active Search~\cite{sattler2016efficient},  typically create the 2D-3D correspondences via structure-from-motion (SfM) reconstruction and achieve state-of-the-art localization accuracy. 
However, the descriptor detector is prone to fail in repetitive or texture-less scenes or complex environments with entangling structures~\cite{shotton2013scene, walch2017imagebased}.

Recently, the learning-based scene coordinate regression methods have been proposed~\cite{brachmann2017dsac, brachmann2018learning, brachmann2019expert, brachmann2019neuralguided, brachmann2021visual, li2020hierarchical, zhou2020kfnet} to predict the 2D-3D matches using neural networks in an end-to-end manner.
Thanks to the great capacity of the neural network, it achieves state-of-the-art performance across various tasks~\cite{brachmann2021visual}.
The regressor learns from the ground-truth scene coordinates and can be theoretically separated from the SfM reconstruction.
% Our method is based on the scene coordinate regression and leverages this property by using the dense and high-quality synthetic coordinate labels created from TOPO-DataGen to avoid the SfM-related issues.
Most existing methods directly learn the mapping from image to coordinates while ignoring the rich geometry information contained in the 3D coordinate labels, $e.g.,$ one could compute camera pose and depth from scene coordinates given camera intrinsics.
\par \noindent
\textbf{Sim-to-real datasets} have become increasingly important to tackle the issue of real data scarcity or to augment the real-world datasets during training.
A set of synthetic datasets have been proposed and employed to facilitate model training for better robustness and efficiency~\cite{chen2020robust, Eftekhar2021Omnidata:Scans, hu2019sailvos, purkait2018synthetic, sankaranarayanan2018learning, sax2020learning, shoman2018illumination, tremblay2018training, Zhao2020DomainEstimation}. 
\cite{Zhao2020DomainEstimation} proposes the attend-remove-complete (ARC) method for sim-to-real depth prediction, which learns to identify, remove and fill in some challenging regions in real images as well as to translate the real images into the synthetic domain. 
% ARC makes good use of synthetic data for accurate depth prediction for real image. 
Closer to our work, \cite{Eftekhar2021Omnidata:Scans} provides a toolkit for mid-level cues generation from 3D models. 
However, it is not optimized for geodata and does not support synthetic data generation to match real geo-tagged photos.
Likewise, most synthetic datasets or data generation workflows only involve rendered results in the virtual world and cannot traverse well between reality and simulation.
Our proposed TOPO-DataGen workflow not only generates multimodal virtual data but can also produce georeferenced reality-matching data to intentionally augment the downstream learning tasks designed by the users.

%------------------------------------------------------------------------
\section{TOPO-DataGen workflow}
In this section, we describe the TOPO-DataGen toolkit for synthetic data generation as shown in Figure~\ref{Figures:main} (left).
% Please refer to our supplementary materials for technical details.
\par \noindent
\textbf{Geodata preprocessing.}
In the first step, a high-fidelity 3D textured model is generated over the area of interest based on available geographic data, $e.g.,$ classified LiDAR point cloud, orthophoto, or digital terrain model.
We preprocess the off-the-shelf geodata such that it is compatible with the open-source geospatial rendering engine CesiumJS~\cite{Cesium3D}, which is at the core of our data generation workflow.
For example, one common practice is to convert the coordinate reference system into the global WGS84 from the local one.
We argue that nowadays, the high-quality open geodata from national agencies is more and more common~\cite{coetzee2020open} such as the swisstopo~\cite{swissimage10cm, swisssurface3D, swisssurface3Draster}, making it easier to access the 3D models for locations of interest and employ our method on a large scale.
Moreover, the open geodata mostly comes from airborne devices such as satellites.
Its accuracy and the top-down view are generally sufficient for aerial localization tasks as opposed to street-level localization.
\par \noindent
\textbf{Synthetic data generation.}
The generated 3D textured model in the WGS84 reference system is the input to CesiumJS engine for synthetic data generation. 
Given the virtual camera viewpoint, our proposed TOPO-DataGen provides a series of designer modalities related to visual localization, which include: RGB image, scene coordinates in the WGS84 reference system ($i.e.,$  earth-centered-earth-fixed coordinates), depth, surface normal, and semantics.
Specifically, through ray tracing~\cite{akenine2019real}, our workflow produces synthetic RGB images and the geo-referenced scene coordinates as raw output.
Subsequently, we retrieve the semantic maps by matching each pixel to its closet point in the categorized geodata, $i.e.$, the classified LiDAR point cloud.
The PyTorch~\cite{paszke2019pytorch} framework is used to accelerate the matrix computation.
Lastly, based on the scene coordinate labels, we generate the other 3D modalities, $i.e.$, depth and surface normal.
Following~\cite{zamir2018taskonomy}, we provide the z-buffer depth, and the surface normal is computed using Open3D~\cite{zhou2018open3d}.
% The Euclidean depth could be obtained from the scene coordinate label and the camera position via L2 norm straightforwardly.
\par
The data rendering can be performed offline over a large area, the size of which is limited by the computer hardware and the available geodata. 
To generate synthetic data from scratch, $e.g.,$ to scan one bounded area, we use the LHS~\cite{Florian1992AnSampling,Manteufel2000EvaluatingSampling} to do minimal yet efficient camera viewpoints sampling.
Otherwise, one may use any real geo-tagged photos to generate the matching synthetic modalities as shown in Figure~\ref{Figures:Fig2} (from the leftmost to the right). 
\par \noindent
\textbf{Quality control.}
The quality of the multimodal synthetic data is essentially dependent on the precision of the 3D model geo-referencing and the accuracy of the image pixel-scene coordinate ray tracing.
Please see the supplement for a detailed case study on the provided benchmark datasets (introduced in Section~\ref{sec:benchmark_dataset}).

\section{Benchmark datasets}
\label{sec:benchmark_dataset}
We introduce two large-scale sim-to-real benchmark datasets to exemplify the utility of the TOPO-DataGen.
% define color
\definecolor{LHS_c}{RGB}{225,161,10}
\definecolor{ip_c}{RGB}{138,226,52}
\definecolor{oop_c}{RGB}{114,159,207}
\begin{figure*}
    \centering
    \includegraphics[height=.18\paperheight]{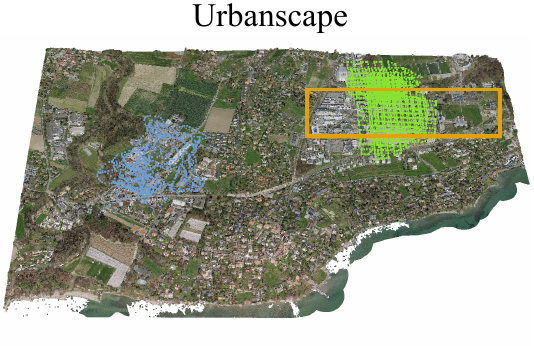}
    % \hfill
    \includegraphics[height=.18\paperheight]{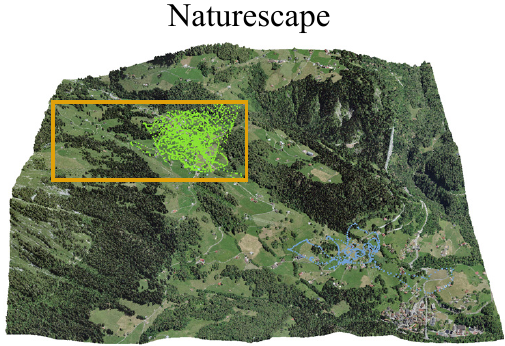}
    % \hfill
    \caption{3D textured models used to render the benchmark datasets with synthetic modalities via LHS sampling or matching the real flight trajectories. 
    Camera positions of \textcolor{ip_c}{in-place} and \textcolor{oop_c}{out-of-place} data are colored in \textcolor{ip_c}{green} and \textcolor{oop_c}{blue} respectively, and the \textcolor{LHS_c}{LHS synthetic} data sampling boundary is denoted by the \textcolor{LHS_c}{orange box}.}
	\label{Figures:Fig3}
\end{figure*}
\begin{table}[tb]
\centering
\small
\scalebox{0.9}{
\begin{tabular}{c|c|ccc}
\toprule
% \multirow{2}{*}{Dataset} & \multirow{2}{*}{Scene}& \multicolumn{3}{c}{Statistics of the benchmark dataset} \\
% \cmidrule(){3-5} 
% & & Size & Image style & Area \\
Landscape & Scene & Size & Image style & Area \\
\midrule
\multirow{3}{*}{\rotatebox[origin=c]{40}{Urbanscape}} & LHS-sim & 15000 & Sim only & 45.8ha\\
& In-place & 3158 & Real-Sim pairs & 40.4ha\\
& Out-of-place & 1360 & Real-Sim pairs & 46.6ha\\
\midrule
\multirow{3}{*}{\rotatebox[origin=c]{40}{Naturescape}} & LHS-sim & 30000 & Sim only & 128.25ha\\
& In-place & 2114 & Real-Sim pairs & 107.5ha\\
& Out-of-place & 565 & Real-Sim pairs & 49.7ha\\
\bottomrule
\end{tabular}}
% \vspace{0.2cm}
\caption{Benchmark datasets statistics.}
\label{tab:dataset_size}
\end{table}
\par \noindent
\textbf{Dataset statistics.}
Table~\ref{tab:dataset_size} shows the statistics of the benchmark datasets, which are distributed across two different landscapes: \textbf{Urbanscape} and \textbf{Naturescape}. 
For each landscape, we provide data in three scenes: \textit{LHS-sim}, \textit{In-place} and \textit{Out-of-place}, all of which come with synthetic 2D images, multimodal 3D labels and semantic map as depicted in Figure~\ref{Figures:Fig2}.
Specially, the \textit{In-place} and \textit{Out-of-place} scenes include accurately geo-tagged real photos captured by a DJI Phantom 4 drone equipped with the cm-level real-time kinematics (RTK) positioning.~\cite{djiphantom4rtk}. 
The \textit{In-place} scene is highly overlapped with the \textit{LHS-sim} scene, while the \textit{Out-of-place} scene describes a neighboring but non-overlapping environment w.r.t \textit{LHS-sim}.
Figure~\ref{Figures:Fig3} shows the 3D textured models with camera position distributions of the datasets.

In the proposed datasets, the data resolution for all modalities, including real and synthetic images, is set to 480 px in height and 720 px in width.
The semantic labels have seven classes: sky, ground, vegetation, building, water, bridge, and others.
Following~\cite{jafarzadeh2021crowddriven}, we evaluate the accuracy of the synthetic 3D labels by coordinate reprojection error.
The mean absolute reprojection error of the Urbanscape and Naturescape datasets are respectively 1.19 px and 1.04 px, showing high accuracy for outdoor aerial datasets. 
More error analysis can be found in the supplement.
\par \noindent
\textbf{Dataset splits.}
We randomly split the \textit{In-place} and \textit{Out-of-place} scene data into training (40\%), validation (10\%) and testing (50\%) sections.
As for the \textit{LHS-sim} scene data, it is split into training (90\%) and validation (10\%) sets.
We intentionally formulate a challenging visual localization task by using more real data for testing than for training to better study the real data scarcity mitigation.
Please also note that the real image density indicated in Table~\ref{tab:dataset_size} is lower than many available outdoor city street-based visual localization datasets such as Cambridge~\cite{kendall2015posenet} and Aachen~\cite{sattler2018benchmarking, sattler2012image}.
\section{CrossLoc localization}
\label{CrossLocArchitecture}
In this section, we present CrossLoc, an absolute localization algorithm leveraging the cross-modal visual representations for enhanced robustness and data efficiency.
CrossLoc learns to localize the query image by predicting its scene coordinates using a set of cross-modal encoders, followed by camera pose estimation using a PnP solver.
\par \noindent
\textbf{Coordinate-depth-normal geometric hierarchy.}
For an image $\mathcal{I}$ with ground-truth scene coordinates $\mathcal{Z}$ and camera pose $(\mathcal{R, P}) \in \bold{SE(3)}$ in the world coordinate system, it is straightforward to compute the Euclidean or z-buffer depth $\mathcal{D}$ via homogeneous transformation.
Subsequently, the surface normal $\mathcal{N}$ can be obtained from the depth map $\mathcal{D}$~\cite{qi2018geonet, zhou2018open3d}.
The geometrical information richness of the scene coordinates $\mathcal{Z}$ label is beyond that of depth $\mathcal{D}$ and surface normal $\mathcal{N}$, as one can convert the first to the latter without loss of accuracy.

We hypothesize that if a neural network is capable of predicting perfect scene coordinates $\mathcal{Z}$, it possesses sufficient information to estimate high quality depth $\mathcal{D}$ and surface normal $\mathcal{N}$.
Inspired by the natural geometric hierarchy, we propose to learn scene coordinate regression with auxiliary self-supervision tasks on depth and surface normal estimation.
Our method is modular and also extends to using any external labels such as semantics $\mathcal{S}$ if they may be available.

Consider an encoder-decoder network making predictions $\mathcal{T}_q \in \{\mathcal{Z}_q, \mathcal{D}_q, \mathcal{N}_q\}$ from a query image $\mathcal{I}_q$.
Let $f^\mathcal{T}(\cdot), g^\mathcal{T}(\cdot), h^{\mathcal{T}}$ denote the encoder, decoder and the mid-way representations respectively:
\begin{equation}
\begin{split}
    h^\mathcal{T}_q &= f^\mathcal{T}(\mathcal{I}_q), \\
    \hat{\mathcal{T}}_q &= g^\mathcal{T}(h^\mathcal{T}_q).
\end{split}
\end{equation}
where $\hat{\mathcal{T}}_q$ is the network prediction.
Based on the coordinate-depth-normal geometric hierarchy, we argue that the representations for scene coordinate $h^{\mathcal{Z}}_q$ may encode richer information than the others, $e.g.,$ $h^{\mathcal{D}}_q, h^{\mathcal{N}}_q$.

To better ensure the hierarchical consistency, we propose to explicitly learn a cross-modal scene coordinate regression representation $h^{\mathcal{Z}}_q$ with those of lower-level tasks:
\begin{equation}
    h^{\mathcal{Z}_+}_q = f^{\mathcal{Z}_+} (h^{\mathcal{Z}}_q, h^{\mathcal{D}}_q, h^{\mathcal{N}}_q).
    \label{equ:feat_merge}
\end{equation}
where $f^{\mathcal{Z}_+}$ is a projection head for representations aggregation.
Next, the cross-modal representation is fed into the primary task decoder for scene coordinate regression:
\begin{equation}
    \hat{\mathcal{Z}}^+_q = g^\mathcal{Z}(h^{\mathcal{Z}_+}_q).
\end{equation}
Afterward, the 6D camera pose could be computed using the coordinates prediction via a PnP solver, as per the standard scene coordinate regression localization methods.
\par \noindent
\textbf{Training objective.}
Let $N$ be the number of pixels in the image.
We use azimuth $\theta$ and elevation $\phi$ to represent the surface normal vectors as in~\cite{li2020deep}, and therefore:
$\mathcal{Z} \in \mathbb{R}^{N\times3}, \mathcal{D} \in \mathbb{R}^{N\times1}, \mathcal{N} = [\mathcal{N}^\theta, \mathcal{N}^\phi] \in \mathbb{R}^{N\times2}$.
The maximum likelihood estimation loss~\cite{kendall2017uncertainties, zhou2020kfnet} is used to stabilize the coordinate, depth and surface normal regression training.
The predicted values come with isotropic noise estimation: $[\hat{\mathcal{T}}, s^\mathcal{T}] = g^\mathcal{T}(h^\mathcal{T})$, where $s^\mathcal{T} \in \mathbb{R}^{N\times1}_{>0}$ is the pixel-wise uncertainty map.
We adopt a two-step training method to regularize the scene coordinate representations $h^{\mathcal{Z}_+}$.
First, the encoder-decoders $f^\mathcal{T}, g^\mathcal{T}$ ($\mathcal{T} \in \{\mathcal{Z}, \mathcal{D}, \mathcal{N}\}$) are separately trained with their own loss functions:
{
\small
\begin{equation}
    \begin{split}
    \small
        L_\mathcal{Z} &= \sum_{i=1}^N \left( \frac{\Vert \hat{\mathcal{Z}}(i) - \mathcal{Z}(i)\Vert^2_2}{2 (s^\mathcal{Z}(i))^2} + 3 \text{ log} (s^\mathcal{Z}(i)) \right), \\
        L_\mathcal{D} &= \sum_{i=1}^N \left( \frac{\Vert \hat{\mathcal{D}}(i) - \mathcal{D}(i)\Vert^2_2}{2 (s^\mathcal{D}(i))^2} + \text{log} (s^\mathcal{D}(i)) \right), \\
        L_\mathcal{N} &= \sum_{i=1}^N \left( \frac{\Vert L_{sn}(\hat{\mathcal{N}}(i), \mathcal{N}(i)) \Vert^2_2}{2 (s^\mathcal{N}(i))^2} + 2\text{ log} (s^\mathcal{N}(i)) \right).
    \end{split}
    \label{equ:crossloc_init_loss}
\end{equation}}
where
{\small
\begin{equation*}
\begin{split}
    L_{sn}(\hat{\mathcal{N}}(i), \mathcal{N}(i)) &= \Vert \hat{\mathcal{N}}^\phi(i) - \mathcal{N}^\phi(i) \Vert_1 + \\ 
     2\text{ min}&\left( \Vert\hat{\mathcal{N}}^\theta(i) - \mathcal{N}^\theta(i)\Vert_1, 1-\Vert\hat{\mathcal{N}}^\theta(i) - \mathcal{N}^\theta(i)\Vert_1 \right)  
\end{split}
\end{equation*}}
is the circle loss defined in~\cite{li2020deep}.
The subscript in Eq.~\eqref{equ:crossloc_init_loss} denotes the loss function for the corresponding task $\mathcal{T}$.
Following~\cite{brachmann2021visual}, we also implement the reprojection loss for coordinate regression loss $L_\mathcal{Z}$.

Lastly, we employ the non-coordinate encoders as frozen representation extractors to fine-tune the coordinate encoder-decoder $f^\mathcal{Z}, g^{\mathcal{Z}}$ and the projection head $f^{\mathcal{Z}_+}$:
\begin{equation}
    L_{crossloc} = L_\mathcal{Z}(f^{\mathcal{Z}}, g^{\mathcal{Z}}, f^{\mathcal{Z}_+}).
    \label{equ:crossloc_final_loss}
\end{equation}
The CrossLoc applies the multi-task learning by re-using the representations of geometrically-related tasks to improve the original objective of coordinate regression. 
Table~\ref{tab:crossloc_variant} shows some variants of the proposed methods, among which the vanilla {CrossLoc} adopts the self-supervised cross-modal representation learning without any additional data.
Our method is flexible with external labels such as semantics, \textit{e.g.,} the CrossLoc-SE.

\begin{table}[t]
\centering
\small
\scalebox{1.0}{
\begin{tabular}{x{2.0cm}|ccc}
\toprule
Encoders & CrossLoc & CrossLoc-SE & CrossLoc-CO\\ 
\midrule
Coordinate & \checkmark & \checkmark & \checkmark\\
Depth & \checkmark & \checkmark &\\
Surface normal & \checkmark & \checkmark &\\
Semantics &  & \checkmark &\\
\bottomrule
\end{tabular}}
\vspace{0.0cm}
\caption{Variants of the proposed CrossLoc algorithm.}
\label{tab:crossloc_variant}
\end{table}

\section{Experiments}
Extensive experiments have been carried out to evaluate the performance of CrossLoc against the state-of-the-art approaches to scene coordinate regression.
Specifically, we apply various CrossLoc architectures in Table~\ref{tab:crossloc_variant} to validate the effectiveness of self-supervision. 
Systematic ablation studies about the efficacy of synthetic training data on real data scarcity mitigation are also conducted.
% We present both quantitative and qualitative results on pose estimation and coordinate regression. 
\par \noindent
\textbf{Network architecture.}
Following~\cite{brachmann2021visual}, we adopt a fully convolutional network (FCN) with ResNet skip layers~\cite{he2016deep} for regression tasks using a downsampling factor of 8.
% Group normalization~\cite{wu2018group} is used at each layer for better convergence.
We split the FCN in the middle to obtain an encoder-decoder structure. Please refer to the supplement for further details.
\par \noindent
\textbf{Datasets.}
Each model is trained using data from the \textit{LHS-sim} and either \textit{In-place} or \textit{Out-of-place} scene.
We evaluate on the real testing data for each scene and report the results.
% While the proposed CrossLoc factually uses more labels than the scene coordinates during training, the additional labels such as depth map could be generated without any external information as explained in Section~\ref{CrossLocArchitecture}.
\par \noindent
\textbf{Training.}
We first initialize the encoder-decoder networks separately using various training tasks on the \textit{LHS-sim} data; subsequently, each network is fine-tuned with the paired real-sim data on either \textit{In-place} or \textit{Out-of-place} scene.
Each network is independently trained using the loss in Eq.~\eqref{equ:crossloc_init_loss}.
Lastly, we fine-tune the coordinator network with loss in Eq.~\eqref{equ:crossloc_final_loss}.
For CrossLoc-SE, we use the cross-entropy loss for semantic segmentation.
We use the Adam optimizer~\cite{kingma2014adam} and the PyTorch~\cite{paszke2019pytorch} framework for implementation.
\par \noindent
\textbf{Baselines.}
We compare our proposed algorithms against two state-of-the-art scene coordinate regression baselines: DSAC*~\cite{brachmann2021visual} and sim-to-real coordinate regression DDLoc adapted from~\cite{Zhao2020DomainEstimation}.
% DDLoc is our own adaptation of the Domain Decluttering architecture~\cite{Zhao2020DomainEstimation} that trains a scene coordinate predictor instead of employing depth estimation as in the original paper.
We follow the ARC~\cite{Zhao2020DomainEstimation} architecture to train a regressor for coordinates instead of depth with minimum modification; see the supplement for details.
For a fair comparison, all the coordinate regression-based approaches utilize the PnP solver from DSAC*~\cite{brachmann2021visual} to compute camera poses.
Also, our method is compared with AtLoc~\cite{wang2020atloc}, a state-of-the-art absolute pose regression (APR) method.
Please note that AtLoc  does not use any 3D labels and has much less information to learn from during training.
 
\subsection{Results on camera pose estimation}
\label{sec:sub_sec_cam_pose}
\begin{table*}[t]
\centering
\small
\begin{subtable}[t]{\linewidth}
\scalebox{1.0}{
\begin{tabular}{c|ccccc|ccccc}
\toprule
& \multicolumn{5}{c|}{In-place localization accuracy} & \multicolumn{5}{c}{Out-of-place localization accuracy} \\ 
\cmidrule(){2-11} 
Methods & \multicolumn{2}{c}{Median error {$\downarrow$}} & \multicolumn{3}{c|}{Accuracy {$\uparrow$}} & \multicolumn{2}{c}{Median error {$\downarrow$}} & \multicolumn{3}{c}{Accuracy {$\uparrow$}} \\
\cmidrule(){2-11} 
& transl. & rot. & $<$ 5m, 5\textdegree & $<$ 10m, 7\textdegree & $<$ 20m, 10\textdegree & transl. & rot. & $<$ 5m, 5\textdegree & $<$ 10m, 7\textdegree & $<$ 20m, 10\textdegree\\
\midrule
DSAC*~\cite{brachmann2021visual} & 11.6m       & 6.2\textdegree       & 15.4\% & 42.1\% & 64.4\% & 14.9m       & 4.1\textdegree       & 10.3\% & 33.1\% & 59.6\%\\
DDLoc & 10.3m       & 2.3\textdegree       & 24.1\% & 47.2\% & 67.8\%  & 42.1m       & 9.5\textdegree       & 4.1\% & 12.8\% & 26.5\%   \\
AtLoc~\cite{wang2020atloc} & 23.0m       & 1.9\textdegree       & 1.6\% & 11.1\% & 40.6\%  & 45.6m       & 5.3\textdegree       & 0.1\% & 2.4\% & 14.1\%   \\
\midrule
% CrossLoc & \textbf{3.9m}       & \textbf{1.9\textdegree}      & 61.3\% & 84.2\% & 92.7\% & 8.2m       & \textbf{2.3\textdegree}       & 19.9\% & \textbf{62.2\%} & 85.7\%       \\
% CrossLoc-SE   & \textbf{3.9m}       & \textbf{1.9\textdegree}       & \textbf{62.3\%} & \textbf{86.8\%} & \textbf{94.6\%}    & \textbf{8.0m}  & 2.4\textdegree & \textbf{22.8\%} & 61.3\% & \textbf{86.2\%}     \\
% CrossLoc-CO & 8.5m       & 4.2\textdegree       & 24.6\% & 56.1\% & 78.4\% & 15.7m       & 4.6\textdegree   & 5.1\%     & 26.5\% & 60.9\%  \\
CrossLoc & {4.0m}       & {2.1\textdegree}      & 61.1\% & 85.6\% & 93.4\% & 6.0m       & 1.9\textdegree       & 39.1\% & 72.4\% & 87.8\%       \\
CrossLoc-SE   & \textbf{3.9m}       & \textbf{1.9\textdegree}       & \textbf{62.2\%} & \textbf{86.7\%} & \textbf{94.2\%}    & \textbf{5.8m}  & \textbf{1.8\textdegree} & \textbf{40.1\%} & \textbf{73.4\%} & \textbf{89.1\%}     \\
CrossLoc-CO & 7.6m       & 3.7\textdegree       & 27.6\% & 61.3\% & 80.5\% & 7.1m       & 2.2\textdegree   & 31.5\%     & 64.9\% & 82.8\%  \\
\bottomrule
\end{tabular}
}
\caption{Quantitative comparison over Urbanscape dataset.}
\label{tab:vloc_urbanscape}
\end{subtable}
\begin{subtable}[t]{\textwidth}
\vspace{0.3cm}
\scalebox{1.0}{
\begin{tabular}{c|ccccc|ccccc}
\toprule
& \multicolumn{5}{c|}{In-place localization accuracy} & \multicolumn{5}{c}{Out-of-place localization accuracy} \\ 
\cmidrule(){2-11} 
Methods & \multicolumn{2}{c}{Median error {$\downarrow$}} & \multicolumn{3}{c|}{Accuracy {$\uparrow$}} & \multicolumn{2}{c}{Median error {$\downarrow$}} & \multicolumn{3}{c}{Accuracy {$\uparrow$}} \\
\cmidrule(){2-11} 
& transl. & rot. & $<$ 5m, 5\textdegree & $<$ 10m, 7\textdegree & $<$ 20m, 10\textdegree & transl. & rot. & $<$ 5m, 5\textdegree & $<$ 10m, 7\textdegree & $<$ 20m, 10\textdegree\\
\midrule
DSAC*~\cite{brachmann2021visual} & 41.9m       & 4.8\textdegree       & 1.8\% & 11.3\%        & 30.1\%        & 33.1m       & 4.3\textdegree        & 0.4\%    & 6.7\% & 26.5\%   \\
DDLoc & 39.8m       & 4.2\textdegree      & 2.2\% & 10.9\%         & 28.0\%       & 57.4m       & 8.2\textdegree       & 2.5\%  & 12.0\% & 22.6\%     \\
AtLoc~\cite{wang2020atloc} & 44.7m       & 5.1\textdegree       & 0.0\% & 1.5\% & 10.7\%  & 70.8m       & 6.6\textdegree       & 0.0\% & 0.0\% & 2.5\%   \\
\midrule
% CrossLoc & \textbf{15.3m}       & \textbf{2.8\textdegree}       & 8.7\% & 31.9\%        & \textbf{59.9\%}      & 19.5m       & 3.4\textdegree & 2.8\% & 17.7\% & 50.9\%      \\
% CrossLoc-SE & \textbf{15.3m}      & 2.9\textdegree     & \textbf{11.7\%} & \textbf{32.4\%}         & 59.3\%        & \textbf{17.6m}       & \textbf{3.2\textdegree}       & \textbf{7.8\%}     & \textbf{26.5\%} & \textbf{55.8\%}  \\
% CrossLoc-CO & 36.7m       & 6.8\textdegree       & 1.5\% & 8.6\%        & 23.9\%       & 45.1m       & 7.4\textdegree       & 0.7\%    & 6.0\%  & 24.4\%  \\
CrossLoc & 16.7m       & 2.9\textdegree       & 9.6\% & 28.4\%        & {59.6\%}      & 15.0m       & {2.7\textdegree} & {8.1\%} & 29.7\% & {59.0\%}      \\
CrossLoc-SE & \textbf{16.0m}      & \textbf{2.7\textdegree}     & \textbf{11.4\%} & \textbf{32.6\%}         & \textbf{59.7\%}        & \textbf{14.5m}       & \textbf{2.2\textdegree}       & \textbf{12.7\%}     & \textbf{32.5\%} & \textbf{59.4\%}  \\
CrossLoc-CO & 22.6m       & 3.8\textdegree       & 6.2\% & 20.1\%        & 42.5\%       & 17.1m       & 2.9\textdegree       & 7.1\%    & 29.0\%  & 55.5\%  \\
\bottomrule
\end{tabular}
}
\caption{Quantitative comparison over Naturescape dataset.}
\label{tab:vloc_naturescape}
\end{subtable}
\caption{
Quantitative results on camera pose estimation.
We report the median translation and rotation errors as well as the percentage of correctly re-localized camera poses below error thresholds of 5m/5\textdegree, 10m/7\textdegree and 20m/10\textdegree.}
\label{tab:vloc_general}
\end{table*}
\begin{figure*}
    \centering
    \includegraphics[width=.99\textwidth]{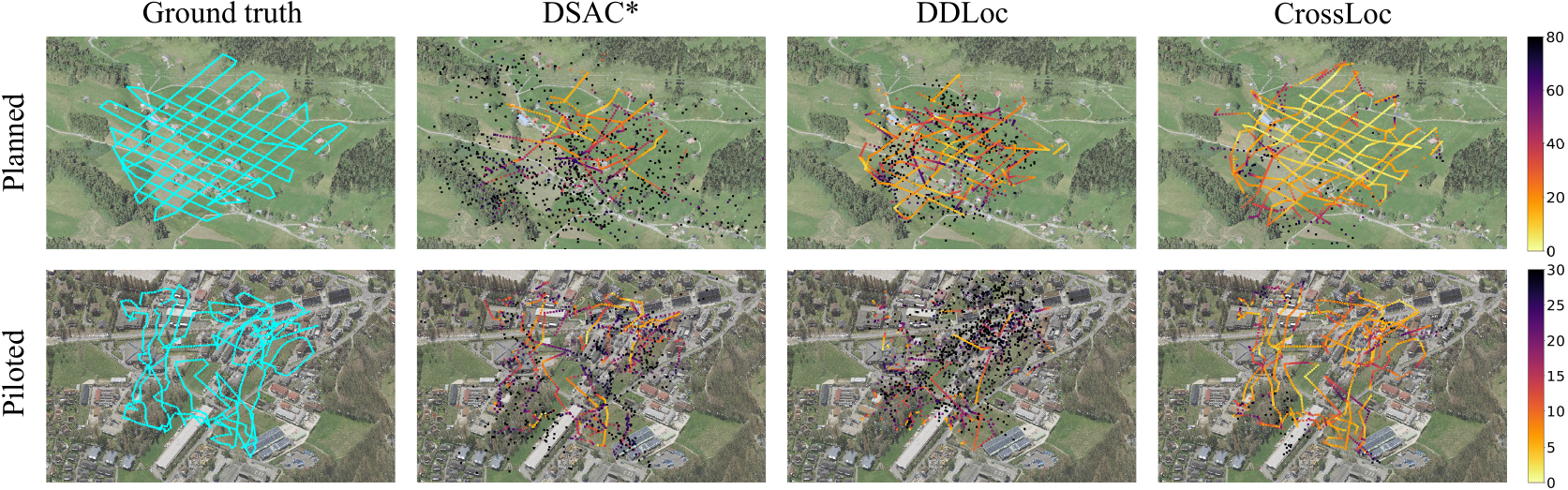}
    \caption{Qualitative comparison of the reconstructed trajectories with the studied visual localization methods.
    We show the results for a planned trajectory on Naturescape and a human-piloted trajectory on Urbanscape, both derived from the testing data.
    % The reconstructed trajectories are based on the estimated camera pose using different localization methods.
    The rightmost color bar denotes the camera position estimation error in meter.
    }
	\label{Figures:Trajectories_accuracy}
\end{figure*}
\vspace{-0.15cm}
\par
\noindent
\textbf{Quantitative results.}
Table~\ref{tab:vloc_general} lists detailed comparisons for pose estimation on {Urbanscape} and {Naturescape} datasets. 
CrossLoc outperforms the other two coordinate regression methods by a clear margin, demonstrating the superior performance of cross-modal representation learning. 
Notably, the 2D image-based APR baseline AtLoc~\cite{wang2020atloc} shows inferior performance to any of the structure-based methods.
This is not unexpected as the APR tends to generalize worse from the training data in practice~\cite{sattler2019understanding}.
We also provide the results of our CrossLoc algorithm using different visual representations, $i.e.,$ CrossLoc-CO applying only coordinate regression features and CrossLoc-SE using additional semantic segmentation features (see Table~\ref{tab:crossloc_variant}).
Adding the auxiliary depth and surface normal visual features substantially boosts the model performance. 
However, CrossLoc-SE with external semantic labels has limited performance gains compared with the vanilla CrossLoc.
This indicates incorporating additional labels with less geometrical information may not help the scene coordinate learning prominently.
This finding is also in line with the recent works on multi-task learning consistency~\cite{qi2018geonet, standley2020tasks, zamir2020robust, zamir2018taskonomy}.
% Additionally, the same set of learning approaches are trained on the \textbf{Naturescape} dataset, and the comparisons are reported in Table~\ref{tab:vloc_naturescape}. 
% We observe similar performance trends as in the Urbanscape dataset. 
% Our CrossLoc method outperforms the two baselines in both in-place and out-of-place scenes. 
% Including the depth and surface normal representations consistently enhances the regression learning, while the performance only changes to a small extent with external semantic labels. 
The localization accuracy in the Naturescape is not as high as in the Urbanscape. 
We conjecture that this is due to the lower amount of human-made features with distinctive geometries such as buildings.
Moreover, the much lower real training data density, as stated in Table~\ref{tab:dataset_size}, makes the localization task considerably harder.
\par \noindent
\textbf{Qualitative results.}
Figure~\ref{Figures:Trajectories_accuracy} shows the pose estimation results of two flight trajectories in the Naturescape and Urbanscape datasets respectively. 
Our CrossLoc outperforms other localization algorithms, resulting in far more complete trajectory reconstruction with much fewer outliers. 
% The two baseline methods result in noisy trajectories with numerous inaccurate pose estimations, while there are almost no such outliers for our method. 

\begin{figure*}
    \centering
    \includegraphics[width=.99\textwidth]{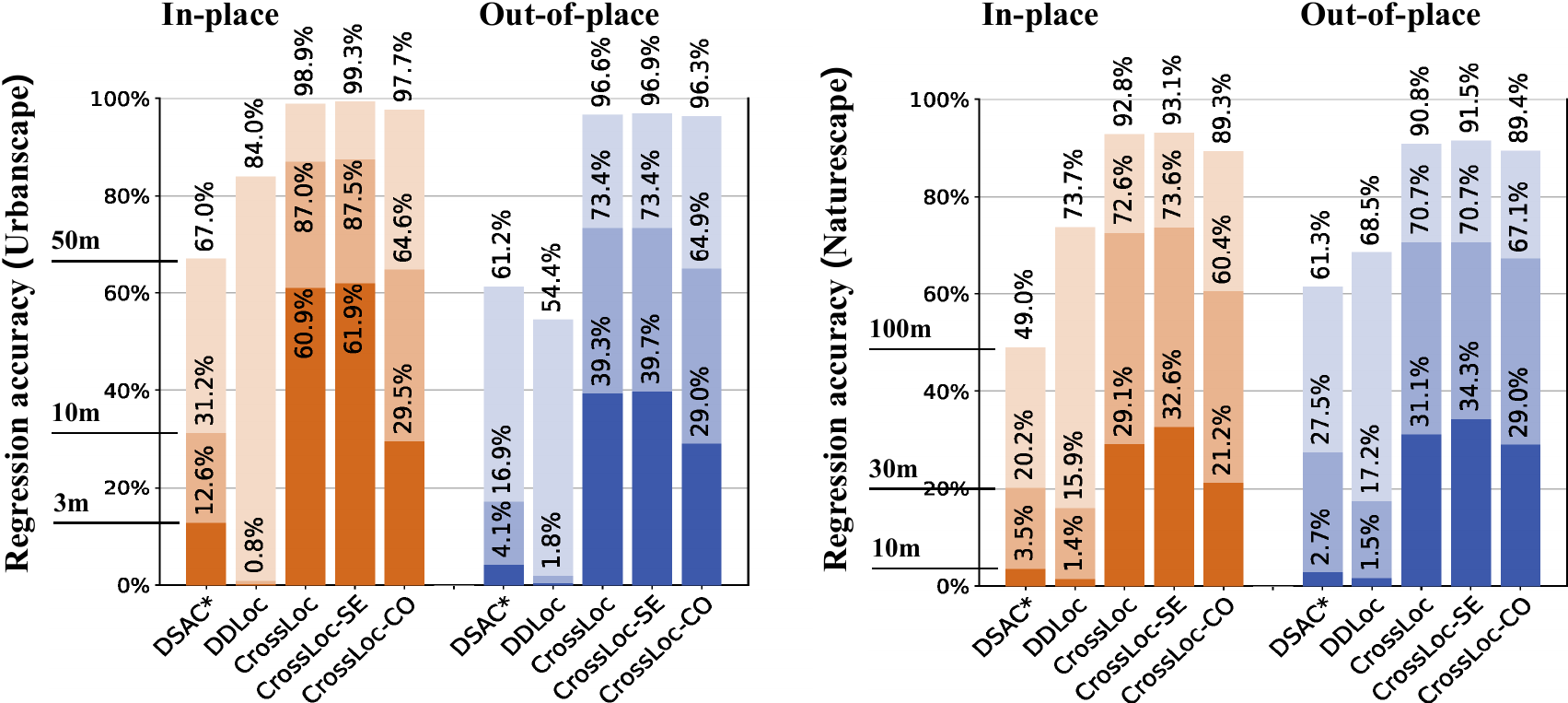}
    \caption{Percentage of correctly predicted scene coordinates with different methods. The accuracy is represented with error thresholds, 3m, 10m and 50m for Urbanscape as well as 10m, 30m and 100m for Naturescape. Only two accuracy values of DDLoc are given in Urbanscape since there is nearly no predicted coordinate with error lower than 3m.}
	\label{Figures:Fig_Scene_coordinates_statistics}
\end{figure*}
\begin{figure*}
    \centering
    \includegraphics[width=.99\textwidth]{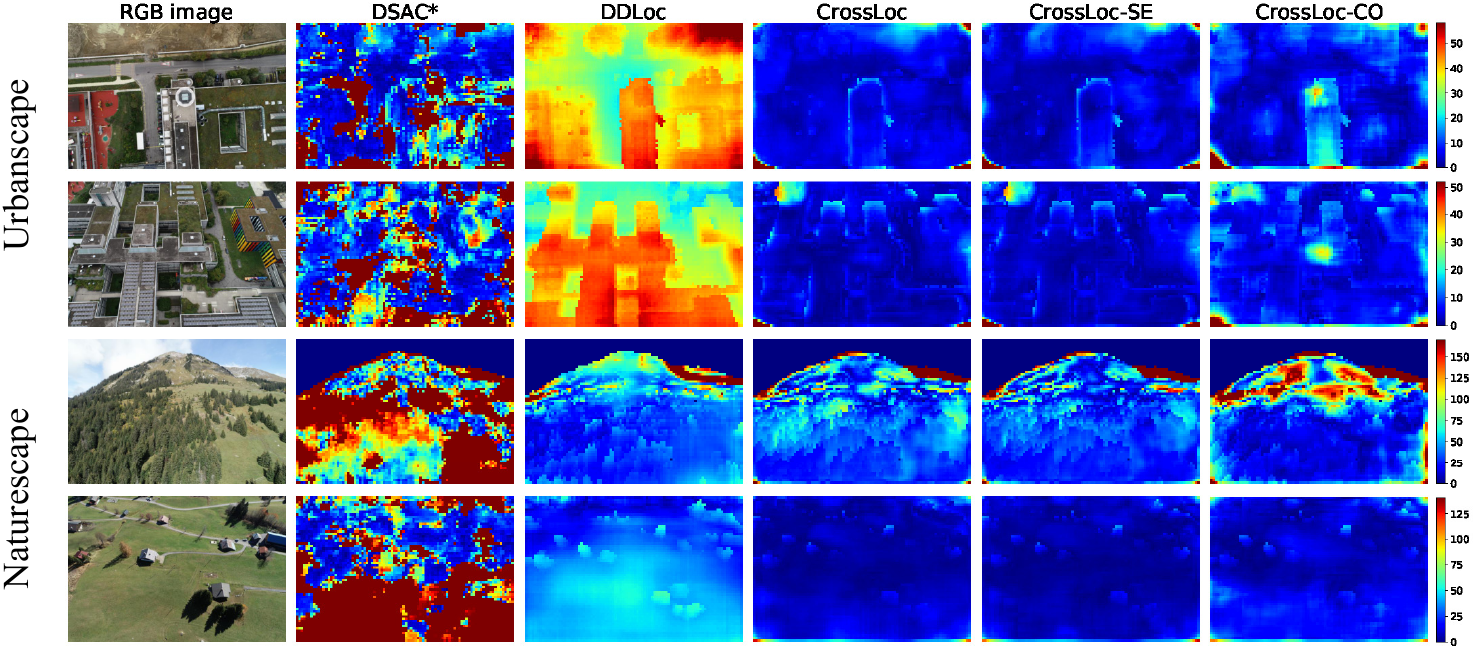}
    \caption{Qualitative comparison of the scene coordinate error map. 
    Pixels without groundtruth labels such as sky are not evaluated.
    Our CrossLoc family methods estimate the absloute scene coordinates with the least error.}
	\label{Figures:Fig_Scene_coordinates_map}
\end{figure*}
\subsection{Results on scene coordinates regression}
\label{sec:sub_sec_coord_regress}
\vspace{-0.15cm}
\par
\noindent
\textbf{Quantitative results.}
For coordinate regression-based localization methods, the final pose estimation performance is highly dependent on the coordinate prediction quality. 
Figure~\ref{Figures:Fig_Scene_coordinates_statistics} shows the accuracy of regressed scene coordinates under different thresholds. 
Consistent with the pose estimation results in Table~\ref{tab:vloc_general}, the CrossLoc architecture frameworks outperform the DSAC* and DDLoc baselines by a substantial margin. 
% Specifically, CrossLoc and CrossLoc-SE achieve around 57\% accuracy with 3m error threshold on the Urbanscape in-place scene.  
% DSAC* and DDLoc, however, fail to reach the same percentage even under the 10m threshold. 
Moreover, CrossLoc and CrossLoc-SE improve the coordinate accuracy over CrossLoc-CO across all scenarios, which further indicates the benefit of leveraging depth and surface normal cross-modal representations. 
% \vspace{0.2cm}
\par
\noindent
\textbf{Qualitative results.}
Figure~\ref{Figures:Fig_Scene_coordinates_map} compares the scene coordinate regression errors of different methods. 
The regression error of our CrossLoc and CrossLoc-SE is significantly lower than the baselines and the plain CrossLoc-CO without self-supervision. 
Further, we observe a high similarity between the error map of CrossLoc and CrossLoc-SE, which shows limited performance gains by using the extra semantic labels. 
For discussion on the failure cases, please refer to the supplementary materials.

\subsection{Ablation study on real data scarcity mitigation}
\label{sec:sub_sec_ablation}
\begin{table}[t]
\centering
\small
\begin{subtable}[t]{\columnwidth}
\scalebox{0.85}{
\begin{tabular}{c|ccc}
\toprule
\multirow{2}{*}{Methods} & \multicolumn{3}{c}{Localization accuracy w.r.t. amount of synthetic data} \\
\cmidrule(){2-4} 
& Real only & Pairwise only & Full \\
\midrule
DSAC*~\cite{brachmann2021visual} & 35.5m, 17.7\textdegree &  23.7m, 12.1\textdegree & 11.6m, 6.2\textdegree\\
DDLoc & - & 11.0m, 2.8\textdegree & 10.3m, 2.3\textdegree\\
\midrule
% CrossLoc & \textbf{7.6m, 3.6\textdegree} & \textbf{4.6m, 2.4\textdegree} & \textbf{3.9m, 1.9\textdegree}\\
% CrossLoc-CO & 14.6m, 7.0\textdegree & 12.8m, 6.2\textdegree & 8.5m, 4.2\textdegree\\
CrossLoc & \textbf{7.6m, 3.6\textdegree} & \textbf{4.6m, 2.4\textdegree} & \textbf{4.0m, 2.1\textdegree}\\
CrossLoc-CO & 14.6m, 7.0\textdegree & 12.8m, 6.2\textdegree & 7.6m, 3.7\textdegree\\
\bottomrule
\end{tabular}}
\caption{Effects of synthetic data assisting the real-world localization.}
\label{tab:ablation_sim_data}
\end{subtable}
\begin{subtable}[t]{\columnwidth}
\vspace{0.3cm}
\scalebox{0.85}{
\begin{tabular}{c|ccc}
\toprule
\multirow{2}{*}{Methods} & \multicolumn{3}{c}{Localization accuracy w.r.t. fraction of pairwise data} \\ 
\cmidrule(){2-4} 
& 25\% data & 50\% data & Full \\
\midrule
DSAC*~\cite{brachmann2021visual} & 33.1m, 17.6\textdegree & 19.4m, 10.2\textdegree & 11.6m, 6.2\textdegree\\
\midrule
% CrossLoc & \textbf{15.6m, 7.8\textdegree} & \textbf{7.7m, 3.7\textdegree} & \textbf{3.9m, 1.9\textdegree} \\
% CrossLoc-CO & 25.0m, 12.4\textdegree & 13.2m, 6.5\textdegree & 8.5m, 4.2\textdegree \\
CrossLoc & \textbf{14.1m, 7.1\textdegree} & \textbf{6.8m, 3.5\textdegree} & \textbf{4.0m, 2.1\textdegree} \\
CrossLoc-CO & 28.3m, 13.7\textdegree & 13.7m, 6.3\textdegree & 7.6m, 3.7\textdegree \\
\bottomrule
\end{tabular}}
\caption{Comparison of real data sample efficiency.}
\label{tab:ablation_matching_data}
\end{subtable}
\caption{Ablation study on the real data scarcity mitigation.
Median translation and rotation error is reported.}
\end{table}
We show the quantitative results on how the real data scarcity issue can be mitigated by: (i) using the synthetic data for augmentation, which could be generated at ease using the proposed TOPO-DataGen, (ii) applying the real data sample-efficient CrossLoc algorithm.
All experiments in this section are carried out on Urbanscape in-place scene.

The added value of the multimodal synthetic data is validated in Table~\ref{tab:ablation_sim_data}. 
We compare the performance of various methods trained with real data only, real-sim paired data only, and real-sim paired data plus the plentiful LHS-sim data.
For each method, the best performance is always achieved with the most synthetic data.
% The synthetic data consistently helps the localization algorithms to understand the geometry of the underlying scene better.
CrossLoc consistently performs the best, which indicates the usefulness of the proposed self-supervision via geometric hierarchy. 

In Table~\ref{tab:ablation_matching_data}, we evaluate the real data sample efficiency of different methods, assuming that the paired synthetic data is available.
We randomly sample 25\% and 50\% of the real training data to train each model.
% Each model is trained using 25\%, 50\% and 100\% of the real training data. 
Each algorithm performs better given a larger amount of training data, and in comparison, the CrossLoc leverages the real data most efficiently.
CrossLoc outperforms the 100\%-data-trained DSAC* and all CrossLoc-CO when trained with only 50\% of available data.
The proposed geometrical self-supervision tasks are particularly helpful for learning in the low-data regime.

\section{Conclusion}
% In this work, we present a scalable visual localization system leveraging multimodal synthetic data.
We propose TOPO-DataGen, a scalable workflow to generate as much synthetic data as needed to assist real-world localization.
Large-scale sim-to-real benchmark datasets are additionally provided to exemplify the use of TOPO-DataGen.
Further, we present CrossLoc that learns to predict coordinates and localize via geometric self-supervision.
It significantly outperforms the state-of-the-art baselines in our benchmark, especially in the low-data regime.
We believe that TOPO-DataGen, altogether with CrossLoc, could open up new opportunities for large-scale aerial localization applications in the real world.

\section*{Acknowledgments}
The support of Armasuisse Science and Technology for this research is very much appreciated. 
We thank Théophile Schenker and Alexis Roger for their projects work, which helped set the spark of this research. 
Finally, we would like to thank Dr. Jan Skaloud, Aman Sharma, and Jesse Lahaye for their comments and proofreading of our manuscript.

\appendix
\section*{Appendix}
\section{Benchmark datasets error analysis}
We validate the accuracy of the multimodal synthetic data generated through our TOPO-DataGen and the geo-tag quality of the real images captured by our drone~\cite{djiphantom4rtk} equipped with the real-time kinematics (RTK) level of Global Navigation Satellite System (GNSS) positioning.

% \subsection{TOPO-DataGen workflow fundamentals}
% The proposed TOPO-DataGen workflow adopts CesiumJS~\cite{Cesium3D} as a core module, which through ray tracing~\cite{akenine2019real} produces synthetic RGB images and the geo-referenced scene coordinate labels corresponding to given camera poses as raw output.
% Subsequently, we retrieve the semantic maps by matching each pixel to its closet point in the categorized geodata, $i.e.$, the classified LiDAR point cloud.
% The PyTorch~\cite{paszke2019pytorch} framework is used to accelerate the matrix computation.
% Lastly, based on the scene coordinate labels, we generate the other 3D modalities, $i.e.$, depth and surface normal.
% Following~\cite{zamir2018taskonomy}, we provide the z-buffer depth, and the surface normal is computed using Open3D~\cite{zhou2018open3d}.
% The Euclidean depth could be obtained from the scene coordinate label and the camera position via L2 norm straightforwardly.
% In summary, the quality of the multimodal synthetic data is essentially dependent on the precision of the \textit{3D model geo-referencing} and the accuracy of the \textit{image pixel-scene coordinate ray tracing}.

\subsection{Synthetic data quality control}
\noindent
\textbf{Geo-referenced 3D scene model precision.}
For the whole benchmark datasets, the digital surface models (DTM)~\cite{swisssurface3Draster} and the LiDAR point clouds~\cite{swisssurface3D} used for rendering were acquired with planimetric precision of $\pm$ 20 cm and altimetric precision of $\pm$ 10 cm.
We employ the orthophoto assets with a position accuracy of $\pm$ 15 cm and a resolution of 10 cm~\cite{swissimage10cm} to colorize the DTM and point clouds.
The $\pm$ sign denotes the standard deviation w.r.t. the local coordinate reference systems~\cite{swissreferenceframe}. 
In the pre-processing step, we convert the open geodata into the global WGS84 coordinate reference system, and the loss of accuracy therein is negligible.
Please see our source code for more details.
\begin{table}[htb]
    \centering
    \scalebox{1.0}{
    \begin{tabular}{c|cc|cc}
        \toprule
         & \multicolumn{2}{c|}{Urbanscape} & \multicolumn{2}{c}{Naturescape} \\
         &  transl. & rot. & transl. & rot. \\
        \midrule 
         Mean & 0.11m & 0.06\textdegree & 0.23m & 0.06\textdegree \\
         Std. & 0.06m & 0.03\textdegree & 0.09m & 0.03\textdegree \\
         Median & 0.10m & 0.06\textdegree & 0.22m & 0.06\textdegree \\
         \bottomrule
    \end{tabular}
    }
    \caption{Indirect quality estimation of the scene coordinate labels. 
    We feed the generated coordinate maps into the DSAC*~\cite{brachmann2021visual} PnP solver and compute the poses error w.r.t. the ground truth camera viewpoints.}
    \label{tab:scene_coord_accuracy}
\end{table}
\begin{table}[htb]
\centering
\scalebox{1.0}{
\centering
    \begin{tabular}{c|c|c}
    \toprule
      & Urbanscape & Naturescape \\
    \midrule
    Mean & 1.19px & 1.04px \\
    Std. & 0.16px & 0.11px\\
    \midrule
    Median & 1.14px & 1.03px\\
    90\% & 1.44px & 1.08px\\
    95\% & 1.52px & 1.09px\\
    99\% & 1.68px & 1.14px\\
    \bottomrule
    \end{tabular}
}
\caption{Absolute reprojection error of the scene coordinate labels. Following~\cite{jafarzadeh2021crowddriven}, we report the reprojection error in the image plane in pixels and particularly show the percentile errors to indicate the long tail distribution.} 
\label{tab:reproj_error}
\end{table}
\par \noindent
\textbf{Scene coordinate ray tracing accuracy.}
The ray-traced scene coordinate label accuracy is further evaluated by verifying the camera pose computation.
Table~\ref{tab:scene_coord_accuracy} demonstrates how well the generated coordinate maps correspond to the ground truth virtual camera viewpoints.
The mean camera pose computation errors for the Urbanscape and Naturescape datasets are 0.11m, 0.06\textdegree, and 0.23m, 0.06\textdegree, respectively.
We use the eight-times-downsampled scene coordinates in this evaluation, the same dataset used in the main paper's experiments.
% We only notice a very slight improvement when using the full-sized coordinate labels.

Table~\ref{tab:reproj_error} reports the reprojection error of the scene coordinates w.r.t. to the ground-truth camera viewpoints as in~\cite{jafarzadeh2021crowddriven}.
The proposed benchmark datasets have consistently low error at the magnitude of one or two pixels in both scenes.
The percentile errors are close to the mean or median errors and indicate that there are few occurrences far away from the central distribution.
The full-size coordinate maps (without downsampling) are used in the reprojection error analysis.

% \begin{figure*}
%     \centering
%     \includegraphics[width=.7\textwidth]{Figures/visualization/Fig_accuracy_ray_tracing_best_fit.png}
%     \caption{Ray tracing point cloud VS Ground truth point cloud}	
%     \label{fig:Accuracy_best_fit}
% \end{figure*}

%  The ray-traced point cloud accuracy was initially estimated by best fitting a complete scene point-cloud (generated by the LHS sampling scene coordinates stiching) to the ground truth 3D model point cloud. The error was estimated with the help of the Cloud Compare open source software. The estimated errors were really small in the order of mm. Example of such study for Urbanscape sceen can be seen in Figure \ref{fig:Accuracy_best_fit}

\subsection{Real data quality control}
\begin{figure*}
    \centering
     \begin{subfigure}[b]{0.33\linewidth}
        \centering
        \includegraphics[height=4cm]{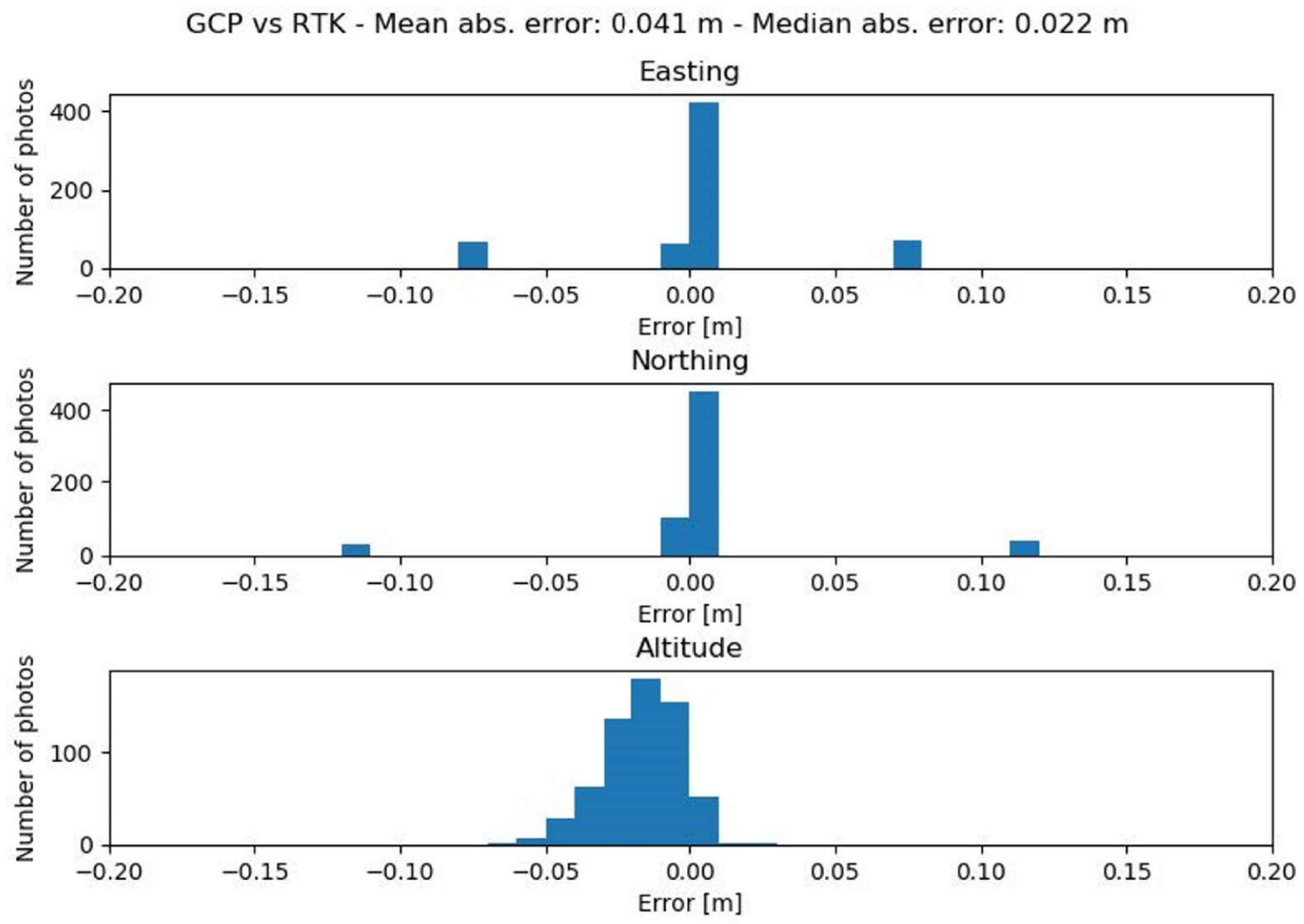}
        \caption{Urbanscape RTK precision: position error.}
        \label{fig:rtk_accuracy_urbanscape}
    \end{subfigure}
     \begin{subfigure}[b]{0.66\linewidth}
     \centering
     \includegraphics[height=4cm]{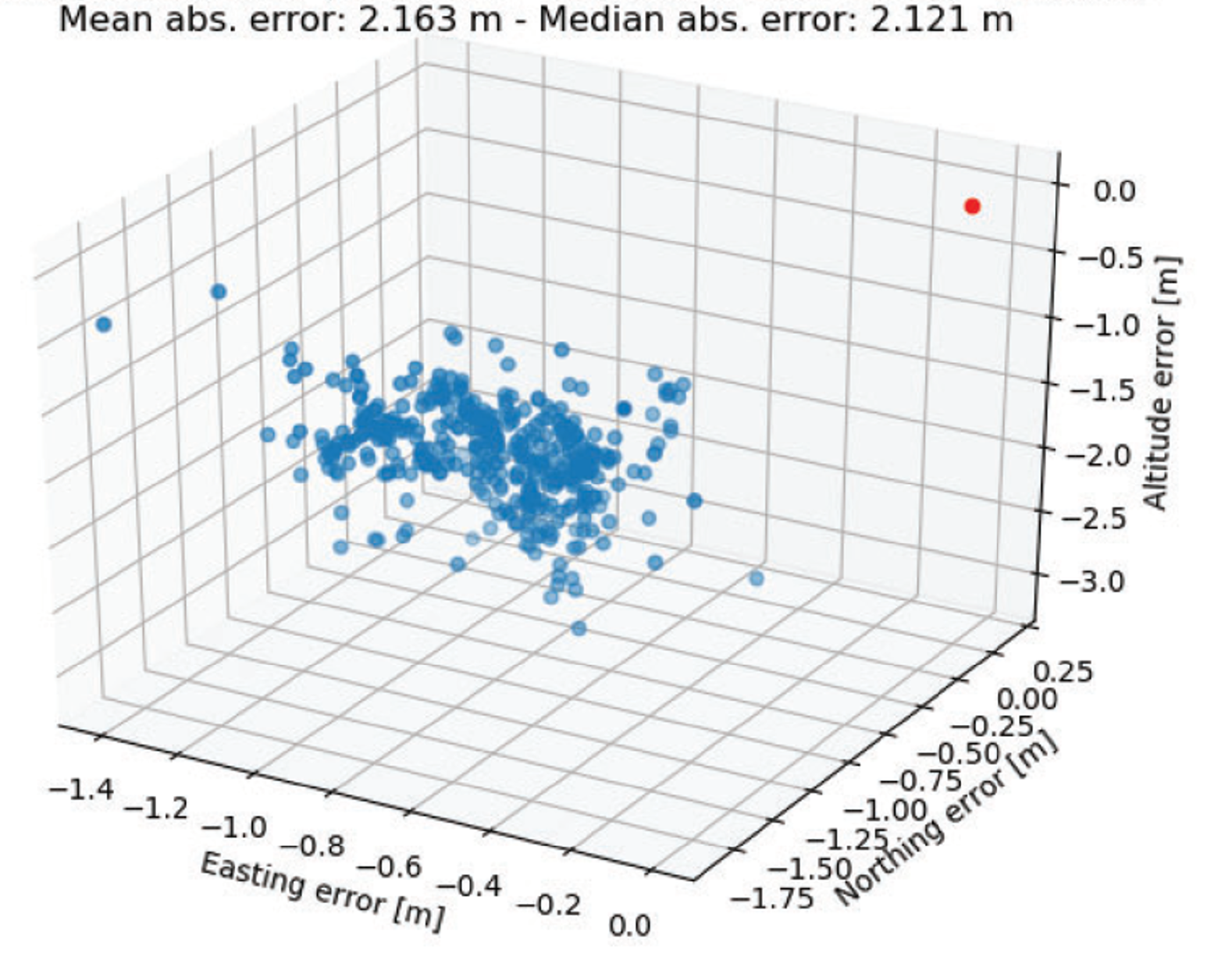}
     \includegraphics[height=4cm]{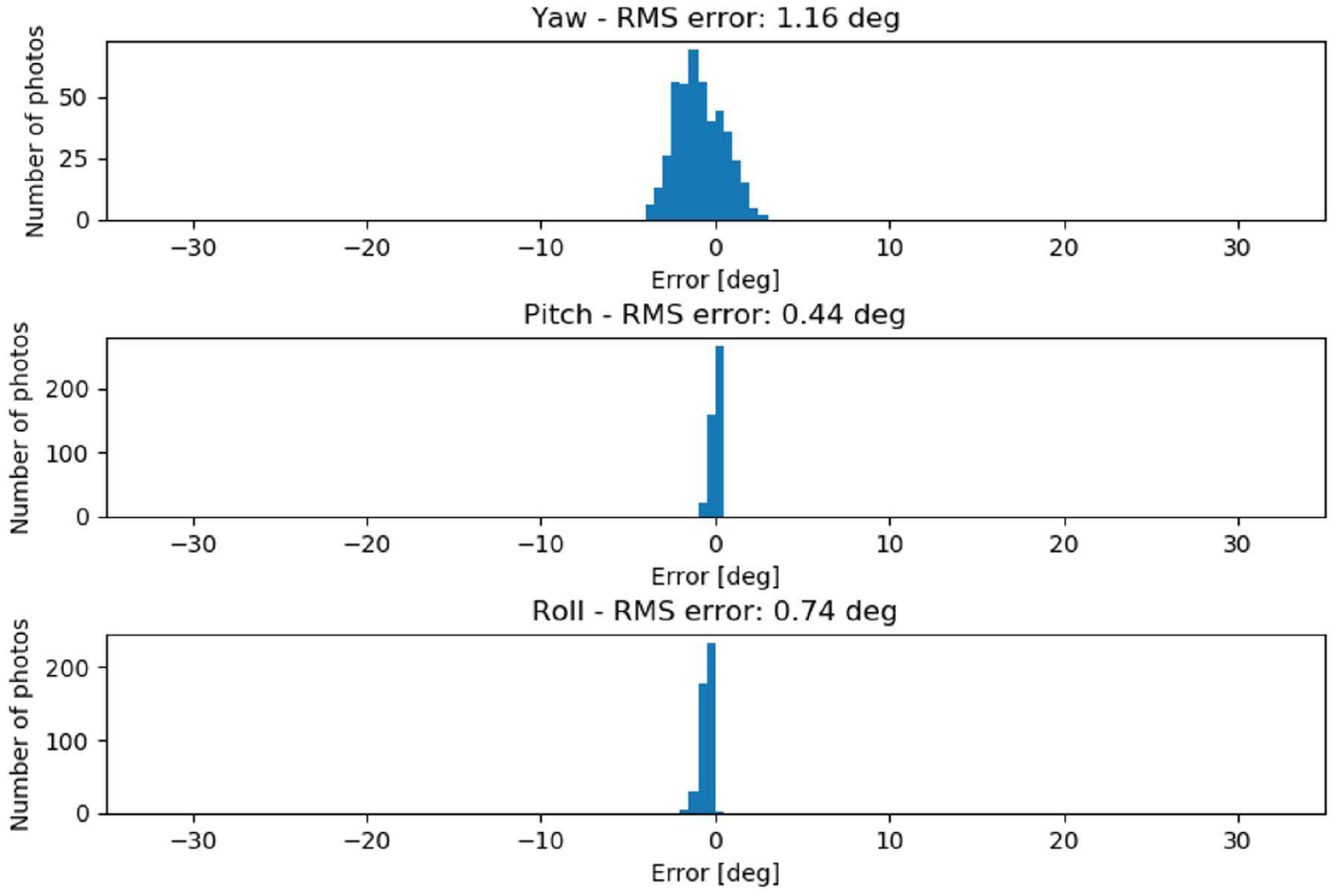}
     \caption{Naturescape RTK precision. \textbf{Left.} Position error. \textbf{Right.} Orientation error.}
     \label{fig:rtk_accuracy_naturescape}
    \end{subfigure}
    \caption{Accuracy evaluation of the real data RTK geo-tag.
    We show the camera pose refinement error obtained by solving the photogrammetry bundle adjustment via GCP alignment.
    The RTK precision in the Urbanscape dataset is at cm level, while it degrades to over 2 meters in the Naturescape dataset because of weak GNSS signals.
    \vspace{0.5cm}
    }
    \label{fig:rtk_accuracy}
\end{figure*}
To collect high-accuracy aerial photos, a DJI Phantom 4 RTK drone~\cite{djiphantom4rtk} was used, whose RTK positioning enables a cm-level accuracy for the image geo-tags. 
A network of ground control points (GCP) was measured with a JAVAD Triumph-LS~\cite{javadtriumphls} base with mm-level accuracy.
We use 13 and 6 GCPs, respectively, for the Urbanscape and the Naturescape datasets to solve the bundle adjustment photogrammetric alignment and compare with the RTK geo-tags extracted from the photo metadata.
Figure~\ref{fig:rtk_accuracy} shows the RTK geo-tag error statistics w.r.t. the computed photogrammetry ground truth.
The geo-tag precision at the Urbanscape dataset is as small as 4 cm, while it degrades significantly in the Naturescape dataset.One can observe that there is a mean positional error
bias. In the next steps, this bias was subtracted for all images of each drone, bringing the error to a lower level (mean error = 0.42 m).
% Those were used in order to enable Post-Processed Kinematic (PPK) references for a photogrammetry survey flight plan was launched using the DJI Phantom 4 Pro RTK. 
% By solving the bundle adjustment photogrammetric alignment with the use of the GCP's an accurate (mm level on position and sub 5th of a degree in orientation) pose reference was estimated. 
% A snapshot of the statistics of one of the Phantom 4 real data flight surveys error w.r.t. the photogrametry derived ground truth can be seen in  Figure~\ref{fig:rtk_accuracy}.

\section{Potential societal impact}
\par \noindent
\textbf{TOPO-DataGen and benchmark datasets.}
The proposed TOPO-DataGen is a multi-purpose task-agnostic synthetic data generation tool that entails little ethical concerns.
Essentially, it needs real-world geo-data to perform data rendering, most of which is provided by national agencies~\cite{coetzee2020open}.
The incoming researchers adopting our method are advised to pay attention to the privacy or transparency of the underlying geo-data before implementation.
In our benchmark datasets, the open-source swisstopo~\cite{swissimage10cm, swisssurface3D, swisssurface3Draster} geo-data is employed for synthetic data generation, and the real data is collected using a drone.
The produced images and associated labels have airborne perspectives, distinct from those in indoor or urban street scenes.
There is no human object data or personally identifiable information in our datasets.

\par \noindent
\textbf{CrossLoc localization.}
Our CrossLoc is a scene coordinate regression-based localization method and does not impose particular requirements infringing privacy.
It learns to localize the image primarily based on the distinct geometries in the environment, such as mountains or buildings.
It does not require human data and is unlikely to benefit from any.

\section{Implementation details}
\subsection{CrossLoc architecture}
\begin{table*}[t]
\centering
\small
\scalebox{0.85}{
\begin{tabular}{c|cccccc|cccccc}
\toprule
& \multicolumn{6}{c|}{Urbanscape} & \multicolumn{6}{c}{Naturescape} \\ 
\cmidrule(){2-13} 
Task & \multicolumn{2}{c}{Encoder pretraining} & \multicolumn{2}{c}{Encoder finetuning} &  \multicolumn{2}{c|}{Encoder finetuning} & \multicolumn{2}{c}{Encoder pretraining} & \multicolumn{2}{c}{Encoder finetuning} & \multicolumn{2}{c}{Encoder finetuning } \\ 
& \multicolumn{2}{c}{ } & \multicolumn{2}{c}{with in-place data} &  \multicolumn{2}{c|}{with out-of-place data} & & &\multicolumn{2}{c}{with in-place data} & \multicolumn{2}{c}{with out-of-place data}\\
\cmidrule(){2-13} 
 & Epoch & Initial LR & Epoch & Initial LR & Epoch & Initial LR & Epoch & Initial LR & Epoch & Initial LR & Epoch & Initial LR \\
\midrule
Coordinate & 150 & 0.0002 & 150 & 0.0002 & 1500 & 0.0002 & 100 & 0.0002  & 150 & 0.0002 & 2000 & 0.0002 \\ 
Depth & 150 & 0.0002 & 150 & 0.0002  & 300 & 0.0002 & 100 & 0.0002  & 150 & 0.0002 & 2000 & 0.0002 \\ 
Surface normal  & 150 & 0.0002 & 150 & 0.0002  & 300 & 0.0002 & 100 & 0.0002  & 150 & 0.0002 & 2000 & 0.0002 \\ 
Semantics  & 30 & 0.0002 & 30 & 0.0002 & 30  & 0.0002 & 30 & 0.0002  & 30 & 0.0002 & 30 & 0.0002 \\
\bottomrule
\end{tabular}
}
% \vspace{0.2cm}
\caption{CrossLoc encoder-decoder initialization training hyper-parameters.}
\label{tab:crossloc_training_details}
\end{table*}

\begin{table}
    \centering
    \begin{subtable}{\columnwidth}
    \centering
    \scalebox{0.7}{
    \begin{tabular}{c|c|c|c}
    \toprule
    Layer & Channel I/O & Kernel/Str./Pad. & Input \\
    \midrule
    conv1 & 3/32 & 3/1/1 & image \\
    conv2 & 32/64 & 3/2/1 & conv1 \\
    conv3 & 64/128 & 3/2/1 & conv2 \\
    conv4 & 128/256 & 3/2/1 & conv3 \\
    \midrule
    res1\_conv1 & 256/256 & 3/1/1 & conv4 \\
    res1\_conv2 & 256/256 & 1/1/0 & res1\_conv1 \\
    res1\_conv3 & 256/256 & 3/1/1 & res1\_conv2 \\
    \midrule
    res2\_add & -/- & -/-/- & \texttt{relu}(res1\_conv3+conv4) \\
    res2\_conv1 & 256/512 & 3/1/1 & res2\_add \\
    res2\_conv2 & 512/512 & 1/1/0 & res2\_conv1 \\
    res2\_conv3 & 512/512 & 3/1/1 & res2\_conv2 \\
    res2\_conv\_s & 256/512 & 1/1/0 & res2\_add \\
    \midrule
    res3\_add & -/- & -/-/- & \texttt{relu}(res2\_conv3+res2\_conv\_s) \\
    res3\_conv1 & 512/512 & 3/1/1 & res3\_add \\
    res3\_conv2 & 512/512 & 1/1/0 & res3\_conv1 \\
    res3\_conv3 & 512/512 & 3/1/1 & res3\_conv2 \\
    \midrule
    res4\_add & -/- & -/-/- & \texttt{relu}(res3\_conv3+res3\_add) \\
    res4\_conv1 & 512/512 & 3/1/1 & res4\_add \\
    res4\_conv2 & 512/512 & 1/1/0 & res4\_conv1 \\
    res4\_conv3 & 512/512 & 3/1/1 & res4\_conv2 \\
    \midrule
    feat\_enc & -/- & -/-/- & \texttt{relu}(res4\_conv3+res4\_add) \\
    \bottomrule
    \end{tabular}}
    \caption{Encoder architecture.
    \vspace{0.5cm}}
    \label{tab:crossloc_encoder}
    \end{subtable}
    \begin{subtable}{\columnwidth}
    \centering
    \scalebox{0.7}{
    \begin{tabular}{c|c|c|c}
    \toprule
    Layer & Channel I/O & Kernel/Str./Pad. & Input \\
    \midrule
    res1\_conv1 & 512/512 & 3/1/1 & feat\_dec \\
    res1\_conv2 & 512/512 & 1/1/0 & res1\_conv1 \\
    res1\_conv3 & 512/512 & 3/1/1 & res1\_conv2 \\
    \midrule
    res2\_add & -/- & -/-/- & \texttt{relu}(res1\_conv3+feat\_dec) \\
    res2\_conv1 & 512/512 & 3/1/1 & res2\_add \\
    res2\_conv2 & 512/512 & 1/1/0 & res2\_conv1 \\
    res2\_conv3 & 512/512 & 3/1/1 & res2\_conv2 \\
    \midrule
    res3\_add & -/- & -/-/- & \texttt{relu}(res2\_conv3+res2\_add) \\
    res3\_conv1 & 512/512 & 1/1/0 & res3\_add \\
    res3\_conv2 & 512/512 & 1/1/0 & res3\_conv1 \\
    res3\_conv3 & 512/512 & 1/1/0 & res3\_conv2 \\
    \midrule
    fc1 & 512/512 & 1/1/0 & \texttt{relu}(res3\_conv3+res3\_add) \\
    fc2 & 512/512 & 1/1/0 & fc1 \\
    output & 512/$N$ & 1/1/0 & fc2 \\
    \bottomrule
    \end{tabular}}
    \caption{Decoder architecture. \vspace{0.5cm}}
    \label{tab:crossloc_decoder}
    \end{subtable}
    \begin{subtable}{\columnwidth}
    \centering
    \scalebox{0.7}{
    \begin{tabular}{c|c|c|c}
    \toprule
    Layer & Channel I/O & Kernel/Str./Pad. & Input \\
    \midrule
    feat\_add & -/- & -/-/- & feat\_enc\_1 $\bigoplus$ $\cdots$ $\bigoplus$ feat\_enc\_{T} \\
    % \midrule
    feat\_conv1 & 512$T$/512 & 3/1/1 & feat\_add\\
    feat\_conv2 & 512/512 & 1/1/0 & feat\_conv1\\
    feat\_conv3 & 512/512 & 3/1/1 & feat\_conv2\\
    feat\_skip & 512$T$/512 & 1/1/0 & feat\_add\\
    \midrule
    feat\_cat & -/- & -/-/- & \texttt{relu}(feat\_conv3 + feat\_skip) 
    \\
    \bottomrule
    \end{tabular}}
    \caption{Representation projection head architecture.}
    \label{tab:crossloc_projection_head}
    \end{subtable}
    \caption{CrossLoc network architecture.}
    \label{tab:crossloc_network}
\end{table}

\par \noindent
\textbf{Network architecture.}
Following~\cite{brachmann2021visual}, we adopt a fully convolutional network with ResNet-style skip layers~\cite{he2016deep} to employ the diverse data augmentation, including rescaling and rotation.
Table~\ref{tab:crossloc_network} shows the CrossLoc encoder-decoder network and the projection head for representations concatenation.
We apply group normalization~\cite{wu2018group} and $\texttt{relu}$ nonlinear activation function for each convolutional or residual layer.
The parameter $N$ in Table~\ref{tab:crossloc_decoder} is task dependent, $e.g.$, for coordinate regression $N=4$ because of 3-dimensional coordinate prediction and 1-dimensional uncertainty estimation.
The parameter $T$ in Table~\ref{tab:crossloc_projection_head} refers to the number of visual representations; for the vanilla \textit{CrossLoc}, $T=3$, and for the \textit{CrossLoc-SE} using external semantics, $T=4$.
The sign $+$ and $\bigoplus$ respectively stand for addition and channel-wise concatenation operators.
As in~\cite{brachmann2021visual}, we use consecutive three convolutional layers with strides of two to downsample the prediction eight times.
For the semantic segmentation task, we keep the full-size labels and apply the dense upsampling convolution~\cite{wang2018understanding} in the final output layer to recover the full-size semantic prediction.
\par \noindent
\textbf{Training hyper-parameters.}
In the first step of CrossLoc training, we pretrain the sub-task networks with task-agnostic \textit{LHS-sim} synthetic data, and fine-tune the models with pairwise real-sim data.
Table~\ref{tab:crossloc_training_details} shows the specific training hyper-parameters.
The learning rate is halved at 50 and 100 epochs at most twice.
We extend the encoder fine-tuning epochs particularly on the out-of-place scene in both datasets to ensure convergence.
Notably, the semantic segmentation tasks reach convergence much faster than the other regression tasks. 
Subsequently, during coordinate network fine-tuning using the frozen non-coordinate encoders as feature extractors, we train each model for 1000 epochs with a fixed learning rate of 0.0001.
In line with~\cite{shan2021improving, Zhao2020DomainEstimation}, we find that reusing fixed modules instead of training from scratch simultaneously improves training convergence.
The Adam optimizer~\cite{kingma2014adam} is used throughout our training.

% \begin{table*}[t]
% \centering
% \small
% \scalebox{0.9}{
% \begin{tabular}{c|cccccccc}
% \toprule
% \cmidrule(){2-9} 
% Task & \multicolumn{2}{c}{Real data only} & \multicolumn{2}{c}{Pairwise data only} &  \multicolumn{2}{c}{25\% pairwise data finetuning} & \multicolumn{2}{c}{50\% pairwise data finetuning} \\ 
% \cmidrule(){2-9} 
%  & Epoch & Initial LR & Epoch & Initial LR & Epoch & Initial LR & Epoch & Initial LR\\
% \midrule
% Coordinate & 800 & 0.0001 (const.) & 400 & 0.0001 (const.) & 400 & 0.0001 (const.) & 400 & 0.0001 (const.)  \\ 
% Depth & 800 & 0.0001 (const.) & 400 & 0.0001 (const.)  & 400 & 0.0001 (const.) & 400 & 0.0001 (const.) \\ 
% Surface normal  & 800 & 0.0001 (const.) & 400 & 0.0001 (const.)  & 400 & 0.0001 (const.) & 400 & 0.0001 (const.)  \\ 
% % Semantics  & 30 & 0.0002 & 30 & 0.0002 & 30  & 0.0002 & 30 & 0.0002 \\
% \bottomrule
% \end{tabular}
% }
% \vspace{0.2cm}
% \caption{CrossLoc ablation study encoder training hyper-parameters}
% \label{tab:crossloc_ablation_training_details}
% \end{table*}

% \begin{table*}[t]
% \centering
% \small
% \scalebox{0.9}{
% \begin{tabular}{c|cccc}
% \toprule
% \cmidrule(){2-5} 
% \multirow{2}{*}{Methods} & \multicolumn{4}{c}{Urbanscape} \\ 
% \cmidrule(){2-5} 
%  & In $\to$ In & In $\to$ Out & Out $\to$ In & Out $\to$ Out\\
% \midrule
% MIoU & 62.39$\%$ & 40.80$\%$ & 46.28$\%$ & 43.75$\%$\\ 
% fwIoU & 83.73$\%$ & 59.23$\%$ & 69.00$\%$ & 59.83$\%$\\ 
% \bottomrule
% \end{tabular}
% }
% \vspace{0.2cm}
% \caption{CrossLoc semantic segmentation performance in Urbanscape}
% \label{tab:crossloc_ablation_training_details}
% \end{table*}

\subsection{DDLoc architecture}
\noindent
\textbf{ARC structure.}
DDLoc is our adaption of the attend-remove-complete (ARC) framework~\cite{Zhao2020DomainEstimation}, which is a state-of-the-art domain transformation method.
The original ARC architecture contains a style translator mapping images between real-world and synthetic domains.
It further trains an attention module to detect challenging regions and an inpainting module to complete the masked regions with realistic fill-in.
A depth predictor module then takes the translated result as input to make the prediction.
The ARC method has been verified by extensive experiments that it can leverage synthetic data for accurate depth estimations~\cite{Zhao2020DomainEstimation}.
\par \noindent
\textbf{Network architecture.}
In our implementation of DDLoc, the depth predictor is replaced with a coordinate regressor, which is implemented by an encoder-decoder architecture~\cite{zhu2017unpaired} with skip connections~\cite{zheng2018t2net}. Based on that, the scene coordinate prediction is further down-sampled by the factor of 8 to enhance efficiency as well as to increase the receptive field~\cite{brachmann2021visual}. The down-sampling is implemented by fully convolution layers with stride of 2. Any other single module in DDLoc uses the same encoder-decoder architecture as used in~\cite{zhu2017unpaired}. The decoder of the attention module is modified to output a single channel to discover challenging regions from real-world input images.
\par \noindent
\textbf{Training hyper-parameters.}
Following ARC's training method~\cite{Zhao2020DomainEstimation}, we first pretrain each module individually using the Adam optimizer~\cite{kingma2014adam} with an initial learning rate of 0.0001 and coefficients of 0.9 and 0.999.
For training the coordinate regressor, we adapt the original depth loss for the coordinate regression distance, and employ re-projection loss as in DSAC*~\cite{brachmann2021visual}.
The sparsity level $\rho$ is chosen as 0.9 when training the attention module on the Urbanscape and the Naturescape datasets.
Other modules are trained with the same loss as that in the original ARC implementation.
Lastly, we fine-tune the whole framework with the loss of the coordinate predictor pretraining and the same training parameters.

\section{Additional qualitative results and analysis}
\begin{figure*}
    \centering
    \includegraphics[width=.95\textwidth]{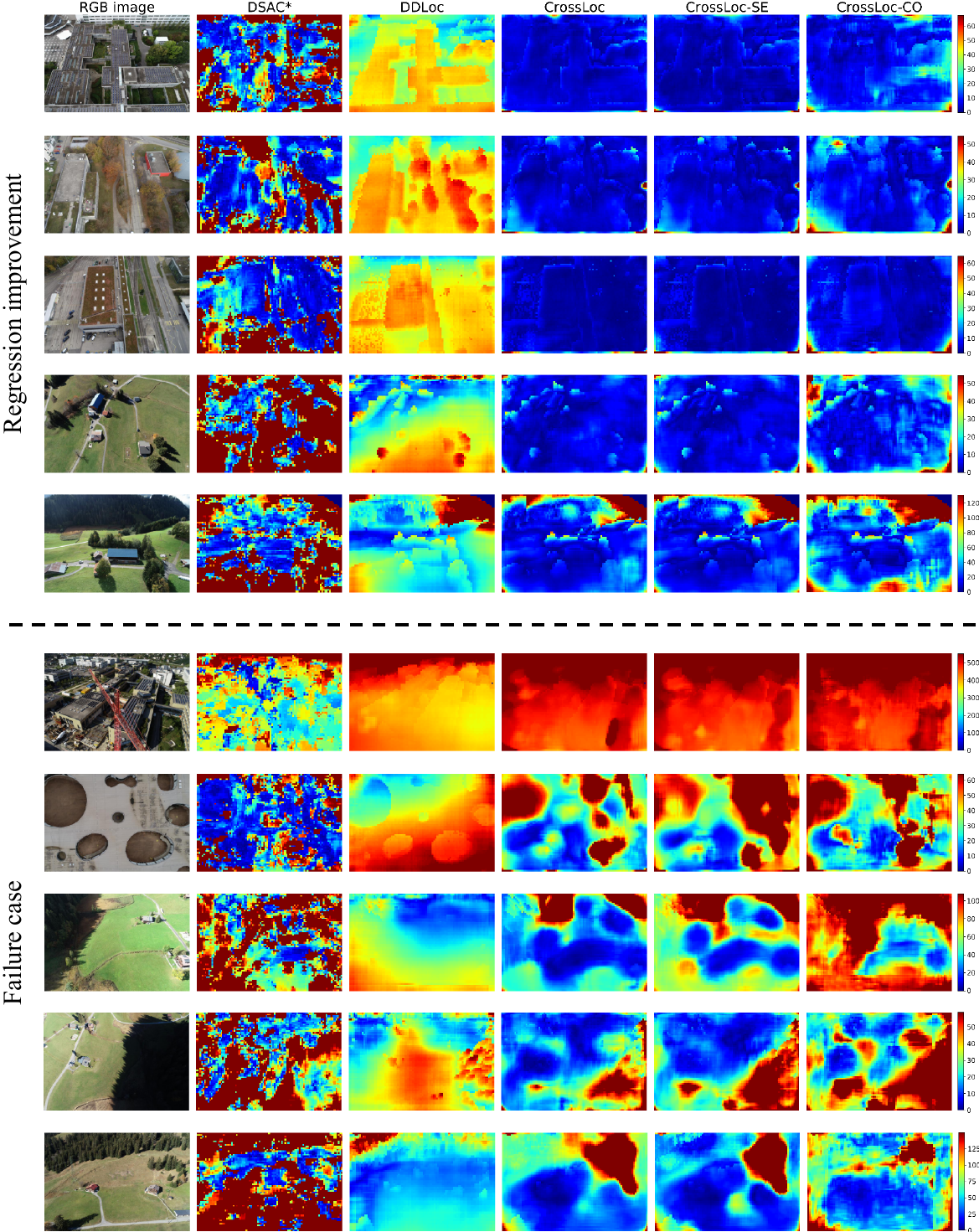}
    \caption{Additional qualitative comparison of the scene coordinate error map including improvement and failure cases. We use the same color bar for visualizing coordinate regression error in each row with the unit in meter.}	
    \label{fig:scene_err_supp}
\end{figure*}

\begin{figure*}
    \centering
    \includegraphics[width=.95\textwidth]{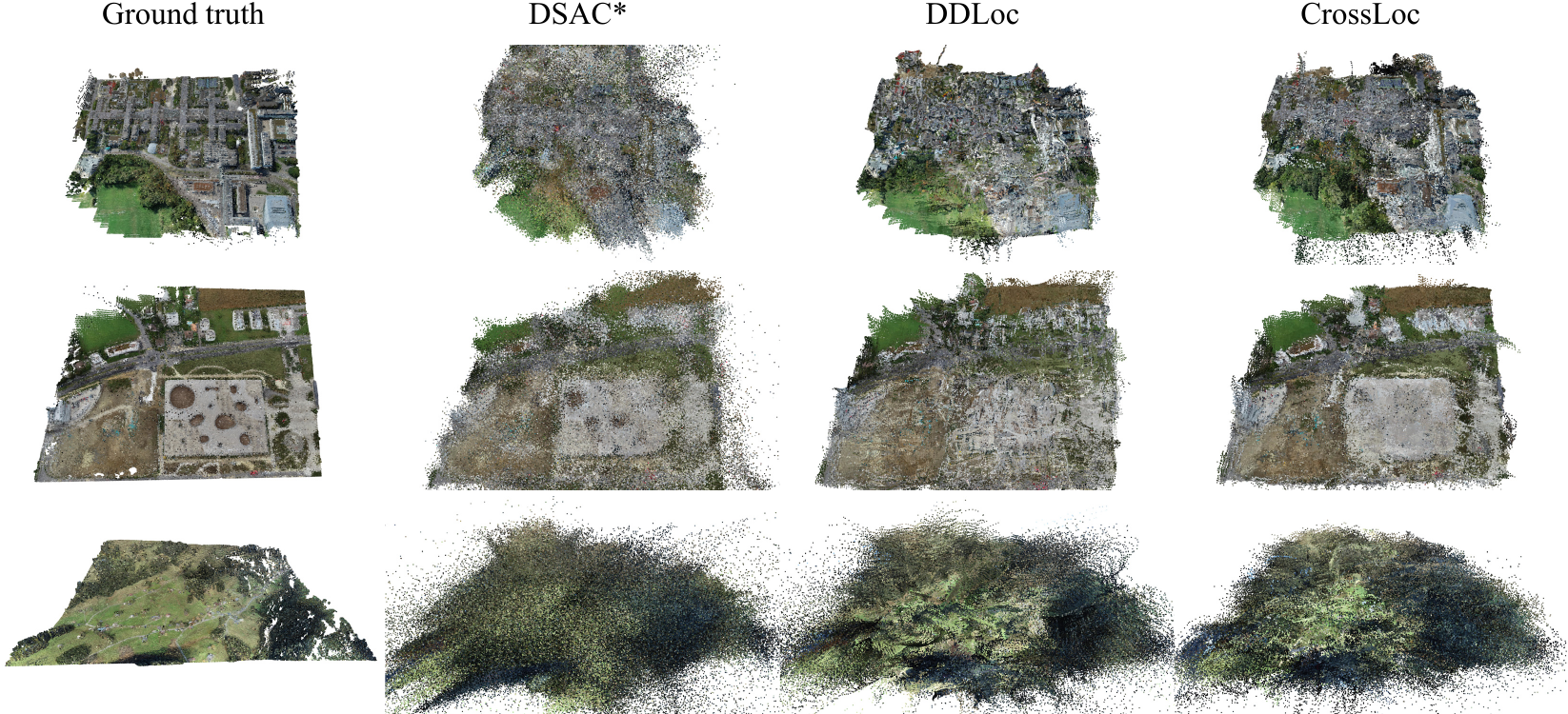}
    \caption{Comparison of point clouds predicted by different coordinate regression approaches. Our CrossLoc method generates a more complete and robust reconstruction of the buildings than the other two baselines.}
    \label{fig:point_cloud}
\end{figure*}
In this section, we visualize additional comparisons of the scene coordinate regression errors in Figure~\ref{fig:scene_err_supp} and compare the predicted point clouds of several urban and natural scenarios in Figure~\ref{fig:point_cloud}.
% In Figure~\ref{fig:scene_err_supp}, additional comparisons of the scene coordinate regression errors are visualized.
We show examples where our CrossLoc outputs accurate coordinate prediction and failure cases where it makes much worse estimation.
Eventually, we summarize the technical limitations of our proposed methods.
\par \noindent
\textbf{Failure cases.}
Generally, CrossLoc and its variants outperform the others by a clear margin. 
Nevertheless, there are two exceptional cases where the CrossLoc family cannot make a good prediction. 
First, novel objects, such as construction cranes, are challenging to the CrossLoc. 
We conjecture that CrossLoc learns the geometric information of the buildings as a whole.
Thus, the appearance of novel objects makes the scene less recognizable for CrossLoc and leads to exacerbated predictions. 
Second, CrossLoc is less robust to the change of illumination conditions. 
The error of CrossLoc prediction increases significantly in light regions and dark shadows in the Naturescape dataset. 
On the contrary, DDLoc predicts the coordinates in these regions more accurately thanks to the translator and attention module.
\par \noindent
\textbf{Point cloud visualization.}
% We present the predicted point clouds of several urban and natural scenario in Figure~\ref{fig:point_cloud}.
As can be observed in Figure~\ref{fig:point_cloud}, CrossLoc gives a complete reconstruction of buildings in the Urbanscape dataset with the fewest outliers.
The predicted point clouds in the Naturescape dataset are generally noisier, which is in line with the quantitative results in the paper.
However, one can still observe that the points predicted by our CrossLoc are less deviated from their actual positions than the other two baselines.
\par \noindent
\textbf{Technical limitations.}
Firstly, the proposed TOPO-DataGen toolkit requires sufficiently-accurate geo-data for training data generation.
Although there are more and more sources of open geo-data nowadays from the national agencies~\cite{coetzee2020open}, it is not likely that they are easily accessible in any location.
The data assets are critical for applying our proposed TOPO-DataGen, which is similar to many other data-driven practices.

Besides, the CrossLoc relies on CNN-extracted visual representations, and like many peer methods~\cite{brachmann2021visual, sarlin2021back}, it could be specific to the local environment or texture.
It is prone to failure for significant outliers or samples unforeseen during the training stage.
The scalability of the proposed CrossLoc may also be limited by the capacity of the CNN backbone network, but it could be addressed by the ensemble regressor learning~\cite{brachmann2019expert} or using more expressive backbone such as the transformers~\cite{dosovitskiy2021an, liu2021swin, vaswani2017attention, shan2022PSTMO}.
% \FloatBarrier
% \newpage
% \clearpage
% {
% \small
% \bibliographystyle{ieee}
% \bibliography{vnav_new}
% }
\clearpage
\newpage
\FloatBarrier

\small
\bibliographystyle{ieee}
\bibliography{vnav_new}

\begin{thebibliography}{10}\itemsep=-1pt

\bibitem{Cesium3D}
{CesiumJS - 3D geospatial visualization for the web}.
\newblock \url{https://cesium.com/platform/cesiumjs/}.

\bibitem{djiphantom4rtk}
{DJI PHANTOM 4 RTK}.
\newblock \url{https://www.dji.com/phantom-4-rtk}.

\bibitem{swissreferenceframe}
{Local Swiss reference frames}.
\newblock
  \url{https://www.swisstopo.admin.ch/en/knowledge-facts/surveying-geodesy/reference-frames/local.html}.

\bibitem{mapillary}
Mapillary.
\newblock \url{https://www.mapillary.com/}.

\bibitem{smapshot}
Smapshot - the participative time machine.
\newblock \url{https://smapshot.heig-vd.ch/?lang=en}.

\bibitem{swissimage10cm}
{SWISSIMAGE 10 cm}.
\newblock
  \url{https://www.swisstopo.admin.ch/en/geodata/images/ortho/swissimage10.html}.

\bibitem{swisssurface3D}
{swissSURFACE3D}.
\newblock
  \url{https://www.swisstopo.admin.ch/en/geodata/height/surface3d.html}.

\bibitem{swisssurface3Draster}
{swissSURFACE3D Raster}.
\newblock
  \url{https://www.swisstopo.admin.ch/en/geodata/height/surface3d-raster.html}.

\bibitem{javadtriumphls}
{TRIUMPH-LS | JAVAD GNSS}.
\newblock \url{https://www.javad.com/jgnss/products/receivers/triumph-ls.html}.

\bibitem{akenine2019real}
T.~Akenine-Moller, E.~Haines, and N.~Hoffman.
\newblock {\em Real-time rendering}.
\newblock AK Peters/crc Press, 2019.

\bibitem{arandjelovic2016netvlad}
R.~Arandjelovic, P.~Gronat, A.~Torii, T.~Pajdla, and J.~Sivic.
\newblock Netvlad: Cnn architecture for weakly supervised place recognition.
\newblock In {\em Proceedings of the IEEE conference on computer vision and
  pattern recognition}, pages 5297--5307, 2016.

\bibitem{arandjelovic2014dislocation}
R.~Arandjelovi{\'c} and A.~Zisserman.
\newblock Dislocation: Scalable descriptor distinctiveness for location
  recognition.
\newblock In {\em Asian Conference on Computer Vision}, pages 188--204.
  Springer, 2014.

\bibitem{brachmann2017dsac}
E.~Brachmann, A.~Krull, S.~Nowozin, J.~Shotton, F.~Michel, S.~Gumhold, and
  C.~Rother.
\newblock Dsac-differentiable ransac for camera localization.
\newblock In {\em Proceedings of the IEEE Conference on Computer Vision and
  Pattern Recognition}, pages 6684--6692, 2017.

\bibitem{brachmann2018learning}
E.~Brachmann and C.~Rother.
\newblock Learning less is more-6d camera localization via 3d surface
  regression.
\newblock In {\em Proceedings of the IEEE Conference on Computer Vision and
  Pattern Recognition}, pages 4654--4662, 2018.

\bibitem{brachmann2019expert}
E.~Brachmann and C.~Rother.
\newblock Expert sample consensus applied to camera re-localization.
\newblock In {\em Proceedings of the IEEE/CVF International Conference on
  Computer Vision}, pages 7525--7534, 2019.

\bibitem{brachmann2019neuralguided}
E.~Brachmann and C.~Rother.
\newblock Neural-guided ransac: Learning where to sample model hypotheses.
\newblock In {\em Proceedings of the IEEE/CVF International Conference on
  Computer Vision}, pages 4322--4331, 2019.

\bibitem{brachmann2021visual}
E.~Brachmann and C.~Rother.
\newblock Visual camera re-localization from rgb and rgb-d images using dsac.
\newblock {\em IEEE Transactions on Pattern Analysis and Machine Intelligence},
  2021.

\bibitem{brahmbhatt2018geometryaware}
S.~Brahmbhatt, J.~Gu, K.~Kim, J.~Hays, and J.~Kautz.
\newblock Geometry-aware learning of maps for camera localization.
\newblock In {\em Proceedings of the IEEE Conference on Computer Vision and
  Pattern Recognition}, pages 2616--2625, 2018.

\bibitem{chen2020robust}
B.~Chen, A.~Sax, F.~Lewis, I.~Armeni, S.~Savarese, A.~Zamir, J.~Malik, and
  L.~Pinto.
\newblock Robust policies via mid-level visual representations: An experimental
  study in manipulation and navigation.
\newblock In J.~Kober, F.~Ramos, and C.~Tomlin, editors, {\em Proceedings of
  the 2020 Conference on Robot Learning}, volume 155 of {\em Proceedings of
  Machine Learning Research}, pages 2328--2346. PMLR, 16--18 Nov 2021.

\bibitem{coetzee2020open}
S.~Coetzee, I.~Iv{\'a}nov{\'a}, H.~Mitasova, and M.~A. Brovelli.
\newblock Open geospatial software and data: A review of the current state and
  a perspective into the future.
\newblock {\em ISPRS International Journal of Geo-Information}, 9(2):90, 2020.

\bibitem{dosovitskiy2021an}
A.~Dosovitskiy, L.~Beyer, A.~Kolesnikov, D.~Weissenborn, X.~Zhai,
  T.~Unterthiner, M.~Dehghani, M.~Minderer, G.~Heigold, S.~Gelly, J.~Uszkoreit,
  and N.~Houlsby.
\newblock An image is worth 16x16 words: Transformers for image recognition at
  scale.
\newblock In {\em International Conference on Learning Representations}, 2021.

\bibitem{iordan2022crossloc}
I.~Doytchinov, Q.~Yan, J.~Zheng, S.~Reding, and S.~Li.
\newblock Crossloc benchmark datasets.
\newblock \url{http://datadryad.org/stash/dataset/doi:10.5061/dryad.mgqnk991c},
  2022.

\bibitem{Eftekhar2021Omnidata:Scans}
A.~Eftekhar, A.~Sax, J.~Malik, and A.~Zamir.
\newblock Omnidata: A scalable pipeline for making multi-task mid-level vision
  datasets from 3d scans.
\newblock In {\em Proceedings of the IEEE/CVF International Conference on
  Computer Vision}, pages 10786--10796, 2021.

\bibitem{ellingson2020relative}
G.~Ellingson, K.~Brink, and T.~McLain.
\newblock Relative navigation of fixed-wing aircraft in gps-denied
  environments.
\newblock {\em NAVIGATION, Journal of the Institute of Navigation},
  67(2):255--273, 2020.

\bibitem{Florian1992AnSampling}
A.~Florian.
\newblock An efficient sampling scheme: updated latin hypercube sampling.
\newblock {\em Probabilistic engineering mechanics}, 7(2):123--130, 1992.

\bibitem{gao2003complete}
X.-S. Gao, X.-R. Hou, J.~Tang, and H.-F. Cheng.
\newblock Complete solution classification for the perspective-three-point
  problem.
\newblock {\em IEEE transactions on pattern analysis and machine intelligence},
  25(8):930--943, 2003.

\bibitem{he2016deep}
K.~He, X.~Zhang, S.~Ren, and J.~Sun.
\newblock Deep residual learning for image recognition.
\newblock In {\em Proceedings of the IEEE conference on computer vision and
  pattern recognition}, pages 770--778, 2016.

\bibitem{hu2019sailvos}
Y.-T. Hu, H.-S. Chen, K.~Hui, J.-B. Huang, and A.~G. Schwing.
\newblock Sail-vos: Semantic amodal instance level video object segmentation-a
  synthetic dataset and baselines.
\newblock In {\em Proceedings of the IEEE/CVF Conference on Computer Vision and
  Pattern Recognition}, pages 3105--3115, 2019.

\bibitem{jafarzadeh2021crowddriven}
A.~Jafarzadeh, M.~L. Antequera, P.~Gargallo, Y.~Kuang, C.~Toft, F.~Kahl, and
  T.~Sattler.
\newblock Crowddriven: A new challenging dataset for outdoor visual
  localization.
\newblock In {\em Proceedings of the IEEE/CVF International Conference on
  Computer Vision}, pages 9845--9855, 2021.

\bibitem{kendall2016modelling}
A.~Kendall and R.~Cipolla.
\newblock Modelling uncertainty in deep learning for camera relocalization.
\newblock In {\em 2016 IEEE international conference on Robotics and Automation
  (ICRA)}, pages 4762--4769. IEEE, 2016.

\bibitem{kendall2017uncertainties}
A.~Kendall and Y.~Gal.
\newblock What uncertainties do we need in bayesian deep learning for computer
  vision?
\newblock {\em Advances in neural information processing systems}, 30, 2017.

\bibitem{kendall2015posenet}
A.~Kendall, M.~Grimes, and R.~Cipolla.
\newblock Posenet: A convolutional network for real-time 6-dof camera
  relocalization.
\newblock In {\em Proceedings of the IEEE international conference on computer
  vision}, pages 2938--2946, 2015.

\bibitem{khaghani2018assessment}
M.~Khaghani and J.~Skaloud.
\newblock Assessment of vdm-based autonomous navigation of a uav under
  operational conditions.
\newblock {\em Robotics and Autonomous Systems}, 106:152--164, 2018.

\bibitem{kingma2014adam}
D.~P. Kingma and J.~Ba.
\newblock Adam: {A} method for stochastic optimization.
\newblock In Y.~Bengio and Y.~LeCun, editors, {\em 3rd International Conference
  on Learning Representations, {ICLR} 2015, San Diego, CA, USA, May 7-9, 2015,
  Conference Track Proceedings}, 2015.

\bibitem{lepetit2009epnp}
V.~Lepetit, F.~Moreno-Noguer, and P.~Fua.
\newblock Epnp: An accurate o (n) solution to the pnp problem.
\newblock {\em International journal of computer vision}, 81(2):155, 2009.

\bibitem{li2020deep}
Q.~Li, J.~Guo, Y.~Fei, Q.~Tang, W.~Sun, J.~Zeng, and Y.~Guo.
\newblock Deep surface normal estimation on the 2-sphere with confidence guided
  semantic attention.
\newblock In {\em European Conference on Computer Vision}, pages 734--750.
  Springer, 2020.

\bibitem{li2020hierarchical}
X.~Li, S.~Wang, Y.~Zhao, J.~Verbeek, and J.~Kannala.
\newblock Hierarchical scene coordinate classification and regression for
  visual localization.
\newblock In {\em Proceedings of the IEEE/CVF Conference on Computer Vision and
  Pattern Recognition}, pages 11983--11992, 2020.

\bibitem{liu2021swin}
Z.~Liu, Y.~Lin, Y.~Cao, H.~Hu, Y.~Wei, Z.~Zhang, S.~Lin, and B.~Guo.
\newblock Swin transformer: Hierarchical vision transformer using shifted
  windows.
\newblock In {\em Proceedings of the IEEE/CVF International Conference on
  Computer Vision}, pages 10012--10022, 2021.

\bibitem{Manteufel2000EvaluatingSampling}
R.~Manteufel.
\newblock Evaluating the convergence of latin hypercube sampling.
\newblock In {\em 41st Structures, Structural Dynamics, and Materials
  Conference and Exhibit}, page 1636, 2000.

\bibitem{meyer2020longterm}
J.~Meyer, D.~Rettenmund, and S.~Nebiker.
\newblock Long-term visual localization in large scale urban environments
  exploiting street level imagery.
\newblock {\em ISPRS Annals of the Photogrammetry, Remote Sensing and Spatial
  Information Sciences}, 2:57--63, 2020.

\bibitem{paszke2019pytorch}
A.~Paszke, S.~Gross, F.~Massa, A.~Lerer, J.~Bradbury, G.~Chanan, T.~Killeen,
  Z.~Lin, N.~Gimelshein, L.~Antiga, et~al.
\newblock Pytorch: An imperative style, high-performance deep learning library.
\newblock {\em Advances in neural information processing systems},
  32:8026--8037, 2019.

\bibitem{purkait2018synthetic}
P.~Purkait, C.~Zhao, and C.~Zach.
\newblock Synthetic view generation for absolute pose regression and image
  synthesis.
\newblock In {\em BMVC}, page~69, 2018.

\bibitem{qi2018geonet}
X.~Qi, R.~Liao, Z.~Liu, R.~Urtasun, and J.~Jia.
\newblock Geonet: Geometric neural network for joint depth and surface normal
  estimation.
\newblock In {\em Proceedings of the IEEE Conference on Computer Vision and
  Pattern Recognition}, pages 283--291, 2018.

\bibitem{sankaranarayanan2018learning}
S.~Sankaranarayanan, Y.~Balaji, A.~Jain, S.~N. Lim, and R.~Chellappa.
\newblock Learning from synthetic data: Addressing domain shift for semantic
  segmentation.
\newblock In {\em Proceedings of the IEEE Conference on Computer Vision and
  Pattern Recognition}, pages 3752--3761, 2018.

\bibitem{sarlin2019coarse}
P.-E. Sarlin, C.~Cadena, R.~Siegwart, and M.~Dymczyk.
\newblock From coarse to fine: Robust hierarchical localization at large scale.
\newblock In {\em Proceedings of the IEEE/CVF Conference on Computer Vision and
  Pattern Recognition}, pages 12716--12725, 2019.

\bibitem{sarlin2021back}
P.-E. Sarlin, A.~Unagar, M.~Larsson, H.~Germain, C.~Toft, V.~Larsson,
  M.~Pollefeys, V.~Lepetit, L.~Hammarstrand, F.~Kahl, et~al.
\newblock Back to the feature: Learning robust camera localization from pixels
  to pose.
\newblock In {\em Proceedings of the IEEE/CVF Conference on Computer Vision and
  Pattern Recognition}, pages 3247--3257, 2021.

\bibitem{sattler2016largescale}
T.~Sattler, M.~Havlena, K.~Schindler, and M.~Pollefeys.
\newblock Large-scale location recognition and the geometric burstiness
  problem.
\newblock In {\em Proceedings of the IEEE conference on computer vision and
  pattern recognition}, pages 1582--1590, 2016.

\bibitem{sattler2016efficient}
T.~Sattler, B.~Leibe, and L.~Kobbelt.
\newblock Efficient \& effective prioritized matching for large-scale
  image-based localization.
\newblock {\em IEEE transactions on pattern analysis and machine intelligence},
  39(9):1744--1756, 2016.

\bibitem{sattler2018benchmarking}
T.~Sattler, W.~Maddern, C.~Toft, A.~Torii, L.~Hammarstrand, E.~Stenborg,
  D.~Safari, M.~Okutomi, M.~Pollefeys, J.~Sivic, et~al.
\newblock Benchmarking 6dof outdoor visual localization in changing conditions.
\newblock In {\em Proceedings of the IEEE Conference on Computer Vision and
  Pattern Recognition}, pages 8601--8610, 2018.

\bibitem{sattler2012image}
T.~Sattler, T.~Weyand, B.~Leibe, and L.~Kobbelt.
\newblock Image retrieval for image-based localization revisited.
\newblock In {\em BMVC}, volume~1, page~4, 2012.

\bibitem{sattler2019understanding}
T.~Sattler, Q.~Zhou, M.~Pollefeys, and L.~Leal-Taixe.
\newblock Understanding the limitations of cnn-based absolute camera pose
  regression.
\newblock In {\em Proceedings of the IEEE/CVF conference on computer vision and
  pattern recognition}, pages 3302--3312, 2019.

\bibitem{sax2020learning}
A.~Sax, J.~O. Zhang, B.~Emi, A.~Zamir, S.~Savarese, L.~Guibas, and J.~Malik.
\newblock Learning to navigate using mid-level visual priors.
\newblock In {\em Conference on Robot Learning}, pages 791--812. PMLR, 2020.

\bibitem{shan2022PSTMO}
W.~Shan, Z.~Liu, X.~Zhang, S.~Wang, S.~Ma, and W.~Gao.
\newblock P-stmo: Pre-trained spatial temporal many-to-one model for 3d human
  pose estimation.
\newblock {\em arXiv:2203.07628}, 2022.

\bibitem{shan2021improving}
W.~Shan, H.~Lu, S.~Wang, X.~Zhang, and W.~Gao.
\newblock Improving robustness and accuracy via relative information encoding
  in 3d human pose estimation.
\newblock In {\em Proceedings of the 29th ACM International Conference on
  Multimedia}, pages 3446--3454, 2021.

\bibitem{shoman2018illumination}
S.~Shoman, T.~Mashita, A.~Plopski, P.~Ratsamee, Y.~Uranishi, and H.~Takemura.
\newblock Illumination invariant camera localization using synthetic images.
\newblock In {\em 2018 IEEE International Symposium on Mixed and Augmented
  Reality Adjunct (ISMAR-Adjunct)}, pages 143--144. IEEE, 2018.

\bibitem{shotton2013scene}
J.~Shotton, B.~Glocker, C.~Zach, S.~Izadi, A.~Criminisi, and A.~Fitzgibbon.
\newblock Scene coordinate regression forests for camera relocalization in
  rgb-d images.
\newblock In {\em Proceedings of the IEEE Conference on Computer Vision and
  Pattern Recognition}, pages 2930--2937, 2013.

\bibitem{standley2020tasks}
T.~Standley, A.~Zamir, D.~Chen, L.~Guibas, J.~Malik, and S.~Savarese.
\newblock Which tasks should be learned together in multi-task learning?
\newblock In {\em International Conference on Machine Learning}, pages
  9120--9132. PMLR, 2020.

\bibitem{toft2020longterm}
C.~Toft, W.~Maddern, A.~Torii, L.~Hammarstrand, E.~Stenborg, D.~Safari,
  M.~Okutomi, M.~Pollefeys, J.~Sivic, T.~Pajdla, et~al.
\newblock Long-term visual localization revisited.
\newblock {\em IEEE Transactions on Pattern Analysis and Machine Intelligence},
  2020.

\bibitem{torii201524}
A.~Torii, R.~Arandjelovic, J.~Sivic, M.~Okutomi, and T.~Pajdla.
\newblock 24/7 place recognition by view synthesis.
\newblock In {\em Proceedings of the IEEE Conference on Computer Vision and
  Pattern Recognition}, pages 1808--1817, 2015.

\bibitem{torii2019are}
A.~Torii, H.~Taira, J.~Sivic, M.~Pollefeys, M.~Okutomi, T.~Pajdla, and
  T.~Sattler.
\newblock Are large-scale 3d models really necessary for accurate visual
  localization?
\newblock {\em IEEE transactions on pattern analysis and machine intelligence},
  43(3):814--829, 2019.

\bibitem{tremblay2018training}
J.~Tremblay, A.~Prakash, D.~Acuna, M.~Brophy, V.~Jampani, C.~Anil, T.~To,
  E.~Cameracci, S.~Boochoon, and S.~Birchfield.
\newblock Training deep networks with synthetic data: Bridging the reality gap
  by domain randomization.
\newblock In {\em Proceedings of the IEEE conference on computer vision and
  pattern recognition workshops}, pages 969--977, 2018.

\bibitem{vaswani2017attention}
A.~Vaswani, N.~Shazeer, N.~Parmar, J.~Uszkoreit, L.~Jones, A.~N. Gomez,
  {\L}.~Kaiser, and I.~Polosukhin.
\newblock Attention is all you need.
\newblock {\em Advances in neural information processing systems}, 30, 2017.

\bibitem{walch2017imagebased}
F.~Walch, C.~Hazirbas, L.~Leal-Taixe, T.~Sattler, S.~Hilsenbeck, and
  D.~Cremers.
\newblock Image-based localization using lstms for structured feature
  correlation.
\newblock In {\em Proceedings of the IEEE International Conference on Computer
  Vision}, pages 627--637, 2017.

\bibitem{wang2020atloc}
B.~Wang, C.~Chen, C.~X. Lu, P.~Zhao, N.~Trigoni, and A.~Markham.
\newblock Atloc: Attention guided camera localization.
\newblock In {\em Proceedings of the AAAI Conference on Artificial
  Intelligence}, volume~34, pages 10393--10401, 2020.

\bibitem{wang2018understanding}
P.~Wang, P.~Chen, Y.~Yuan, D.~Liu, Z.~Huang, X.~Hou, and G.~Cottrell.
\newblock Understanding convolution for semantic segmentation.
\newblock In {\em 2018 IEEE winter conference on applications of computer
  vision (WACV)}, pages 1451--1460. IEEE, 2018.

\bibitem{wu2018group}
Y.~Wu and K.~He.
\newblock Group normalization.
\newblock In {\em Proceedings of the European conference on computer vision
  (ECCV)}, pages 3--19, 2018.

\bibitem{yang2019sanet}
L.~Yang, Z.~Bai, C.~Tang, H.~Li, Y.~Furukawa, and P.~Tan.
\newblock Sanet: Scene agnostic network for camera localization.
\newblock In {\em Proceedings of the IEEE/CVF International Conference on
  Computer Vision}, pages 42--51, 2019.

\bibitem{zamir2020robust}
A.~R. Zamir, A.~Sax, N.~Cheerla, R.~Suri, Z.~Cao, J.~Malik, and L.~J. Guibas.
\newblock Robust learning through cross-task consistency.
\newblock In {\em Proceedings of the IEEE/CVF Conference on Computer Vision and
  Pattern Recognition}, pages 11197--11206, 2020.

\bibitem{zamir2018taskonomy}
A.~R. Zamir, A.~Sax, W.~Shen, L.~J. Guibas, J.~Malik, and S.~Savarese.
\newblock Taskonomy: Disentangling task transfer learning.
\newblock In {\em Proceedings of the IEEE conference on computer vision and
  pattern recognition}, pages 3712--3722, 2018.

\bibitem{zhang2021reference}
Z.~Zhang, T.~Sattler, and D.~Scaramuzza.
\newblock Reference pose generation for long-term visual localization via
  learned features and view synthesis.
\newblock {\em International Journal of Computer Vision}, 129(4):821--844,
  2021.

\bibitem{Zhao2020DomainEstimation}
Y.~Zhao, S.~Kong, D.~Shin, and C.~Fowlkes.
\newblock Domain decluttering: Simplifying images to mitigate synthetic-real
  domain shift and improve depth estimation.
\newblock In {\em Proceedings of the IEEE/CVF Conference on Computer Vision and
  Pattern Recognition}, pages 3330--3340, 2020.

\bibitem{zheng2018t2net}
C.~Zheng, T.-J. Cham, and J.~Cai.
\newblock T2net: Synthetic-to-realistic translation for solving single-image
  depth estimation tasks.
\newblock In {\em Proceedings of the European Conference on Computer Vision
  (ECCV)}, pages 767--783, 2018.

\bibitem{zhou2020kfnet}
L.~Zhou, Z.~Luo, T.~Shen, J.~Zhang, M.~Zhen, Y.~Yao, T.~Fang, and L.~Quan.
\newblock Kfnet: Learning temporal camera relocalization using kalman
  filtering.
\newblock In {\em Proceedings of the IEEE/CVF Conference on Computer Vision and
  Pattern Recognition}, pages 4919--4928, 2020.

\bibitem{zhou2018open3d}
Q.-Y. Zhou, J.~Park, and V.~Koltun.
\newblock {Open3D}: {A} modern library for {3D} data processing.
\newblock {\em arXiv:1801.09847}, 2018.

\bibitem{zhu2017unpaired}
J.-Y. Zhu, T.~Park, P.~Isola, and A.~A. Efros.
\newblock Unpaired image-to-image translation using cycle-consistent
  adversarial networks.
\newblock In {\em Proceedings of the IEEE international conference on computer
  vision}, pages 2223--2232, 2017.

\end{thebibliography}
\end{document}